\documentclass[lettersize,journal]{IEEEtran}
\usepackage{amsmath,amsfonts}
\usepackage{algorithmic}

\usepackage{cite}
\usepackage{amsmath,amssymb,amsfonts}
\usepackage{algorithmic}
\usepackage{graphicx}
\usepackage{algorithm,algorithmic}
\usepackage{hyperref}
\hypersetup{hidelinks=true}
\usepackage{textcomp}
\usepackage{mathrsfs}  
\usepackage{array}
\usepackage{multirow}
\usepackage{booktabs}

\usepackage{accents} 

\usepackage{mathrsfs}

\usepackage{array}
\usepackage[caption=false,font=normalsize,labelfont=sf,textfont=sf]{subfig}
\usepackage{textcomp}
\usepackage{stfloats}
\usepackage{url}
\usepackage{verbatim}
\usepackage{graphicx}
\def\BibTeX{{\rm B\kern-.05em{\sc i\kern-.025em b}\kern-.08em
    T\kern-.1667em\lower.7ex\hbox{E}\kern-.125emX}}
\usepackage{balance}
\begin{document}
\title{Distributed optimization: designed for federated learning
}
\author{Wenyou Guo, Ting Qu,  Chunrong Pan, and George Q. Huang
\thanks{This work was supported in part by the National Natural Science Foundation of China (NSFC) under Grant 52375498,  and in part by the Fundamental Research Funds for the Central Universities under Grant 21623111. (Corresponding author: Ting Qu)}
\thanks{Wenyou Guo is with School of Management, Jinan University, Guangzhou 510632, China (e-mail: guosir1997@163.com). }
\thanks{Ting Qu is  with Guangdong International Cooperation Base of Science and Technology for GBA Smart Logistics, Jinan University, Zhuhai 519070, China, also with School of Intelligent Systems Science and Engineering, Jinan University, Zhuhai 519070, China,  and also with Institute of Physical Internet, Jinan University, Zhuhai 519070, China  (e-mail: quting@jnu.edu.cn).}
\thanks{Chunrong Pan  is  with School of Mechanical and Electrical Engineering, Jiangxi University of Science and Technology, Ganzhou 341000, China  (e-mail: crpan@jxust.edu.cn).}
\thanks{George Q. Huang  is  with Department of Industrial and Systems Engineering, The Hong Kong Polytechnic University, Hong Kong, China  (e-mail: gq.huang@polyu.edu.hk).}
}


\maketitle

\begin{abstract}
Federated learning (FL), as a distributed collaborative machine learning (ML) framework under privacy-preserving constraints, has garnered increasing research attention in cross-organizational data collaboration scenarios. This paper proposes a class of distributed optimization algorithms based on the augmented Lagrangian technique, designed to accommodate diverse communication topologies in both centralized and decentralized FL settings. Furthermore, we develop multiple termination criteria and parameter update mechanisms to enhance computational efficiency, accompanied by rigorous theoretical guarantees of convergence. By generalizing the augmented Lagrangian relaxation through the incorporation of proximal relaxation and quadratic approximation, our framework systematically recovers a broad of classical unconstrained optimization methods, including proximal algorithm, classic gradient descent, and stochastic gradient descent, among others. Notably, the convergence properties of these methods can be naturally derived within the proposed theoretical framework. 
Numerical experiments demonstrate that the proposed algorithm exhibits strong performance in large-scale settings with significant statistical heterogeneity across clients.
\end{abstract}


\section{Introduction}
\label{sec:1.introduction}
\begin{figure*}[t]
    \includegraphics[width=0.96\linewidth]{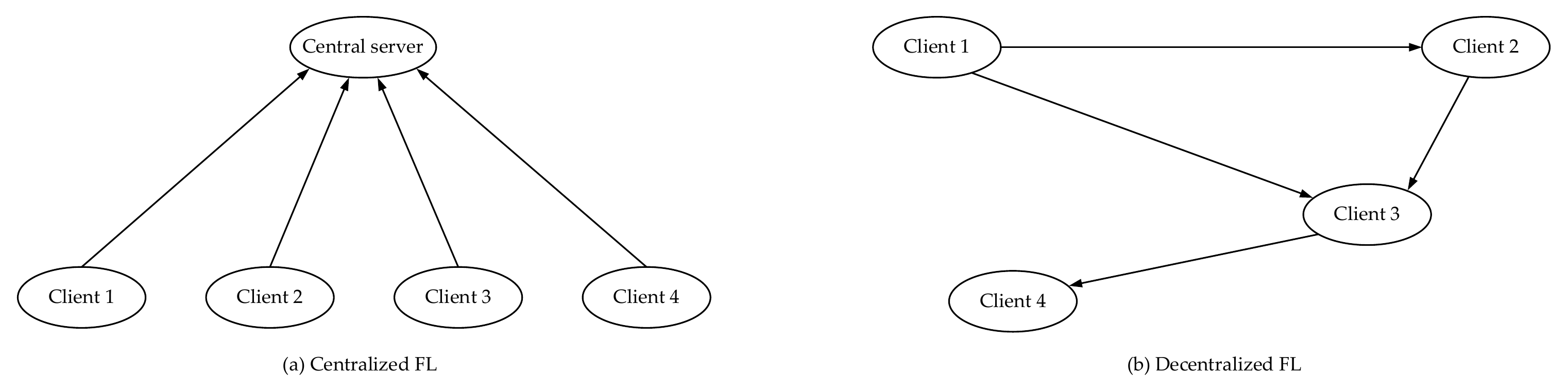}
    \centering
        \caption{Communication Topologies of Centralized and Decentralized Federated Learning}
        \label{fig:1}
\end{figure*}

\IEEEPARstart{I}{n} this paper, we consider a distributed unconstrained optimization problem collaboratively solved by $n$ agents:
\begin{equation}
\min_{\mathbf{x}} f(\mathbf{x}) = \sum_{i=1}^n f_i(\mathbf{x}),
\label{eq:1}
\end{equation}
where each local objective function $f_i(\mathbf{x})$, accessible exclusively to agent $i$, may exhibit non-convexity and non-smoothness, \( i \in \mathbb{N}_n = \{1, 2, \dots, n\} \). Such formulations, commonly referred to as consensus optimization problems, find widespread applications in interdisciplinary domains including distributed ML, collaborative sensing in sensor networks, and distributed parameter estimation~\cite{nedicAchievingGeometricConvergence2017}. These decision-making problems frequently demonstrate inherent large-scale characteristics or involve geographically dispersed data distributions across agents, necessitating distributed processing to mitigate communication overhead while preserving data privacy.

Traditional monolithic optimization frameworks, which rely on centralized information coordination and global data sharing, prove inadequate in addressing these challenges. In contrast, distributed optimization algorithms offer a more scalable and privacy-aware solution through two fundamental mechanisms: 
a) Model decomposition and coordination~\cite{cohenOptimizationDecompositionCoordination1978}: The all-in-one (AIO) model is decoupled into a set of smaller subproblems that can be solved and coordinated across multiple computing nodes, thereby dramatically reducing individual computational burdens.  b) Privacy-preserving computation: Local data processing circumvents raw information exposure through the exchange of essential intermediate parameters (e.g., iterative solutions), maintaining an optimal balance between computational efficiency and data confidentiality.
These merits have established distributed optimization as a pivotal paradigm in large-scale distributed decision-making systems, garnering sustained attention across both academia and industry.

To address problem~(\ref{eq:1}), Nedić \textit{et al}.~\cite{nedicDistributedSubgradientMethods2009,nedicConstrainedConsensusOptimization2010} proposed the Distributed (Sub)Gradient Descent (DGD) method, as defined by the update rule~(\ref{eq:57}). This approach updates each agent's estimate by taking a weighted average with its neighbors, followed by a (sub)gradient descent step based on its local objective function. This idea traces back to the early work of Tsitsiklis \textit{et al}.~\cite{tsitsiklisDistributedAsynchronousDeterministic1986}. Wei \textit{et al}.~\cite{berahasBalancingCommunicationComputation2018,mansooriFastDistributedAsynchronous2020} later interpreted this update rule as equivalent to solving a distributed proximal optimization problem, as formulated in~(\ref{eq:53}). However, they did not provide further theoretical derivations to support this equivalence. Owing to its structural simplicity and ease of implementation, the DGD framework has since been extended to accommodate increasingly complex distributed decision-making scenarios. These extensions include adaptations to time-varying networks~\cite{nedicDistributedOptimizationTimevarying2015,liangDualAveragingPush2020}, non-convex objective functions~\cite{tatarenkoNonconvexDistributedOptimization2017,zengNonconvexDecentralizedGradient2018,wangDecentralizedNonconvexOptimization2023}, constant step sizes~\cite{shiEXTRAExactFirstorder2014,leiDistributedVariableSamplesize2022}, Nesterov-type acceleration~\cite{jakoveticConvergenceRatesDistributed2014,jakoveticFastDistributedGradient2014,quAcceleratedDistributedNesterov2020}, continuous-time system modeling~\cite{linDistributedContinuoustimeDiscretetime2019,linDistributedContinuoustimeOptimization2017,liDistributedAdaptiveConvex2018}, online optimization~\cite{shahrampourDistributedOnlineOptimization2018,luOnlineDistributedOptimization2022,luOnlineDistributedOptimization2023}, and differentially private mechanisms~\cite{dingDifferentiallyPrivateDistributed2022,xuanGradienttrackingBasedDifferentially2023,wangDifferentiallyPrivateDistributed2024,wangTailoringGradientMethods2024}.

Another class of methods for solving problem~(\ref{eq:1}) is based on the Alternating Direction Method of Multipliers (ADMM). The core idea is to introduce a consensus constraint into the primal problem and perform variable updates within the ADMM framework. Representative works in this direction include~\cite{shiLinearConvergenceADMM2014,lingDLMDecentralizedLinearized2015,changMultiagentDistributedOptimization2015,makhdoumiConvergenceRateDistributed2017,caoDifferentiallyPrivateADMM2021,bastianelloAsynchronousDistributedOptimization2021,zhouFederatedLearningInexact2023,zhouFedGiAEfficientHybrid2023,kantFederatedLearningUsing2023}.
It is important to note that, despite the decentralized characteristics exhibited by some of these methods in their formulations, all these works are, in essence, still limited by the classical two-block structure of ADMM. More precisely, they can be regarded as adopting a two-level coordination scheme governed by a central coordinator. As illustrated by the communication topology in Fig.~\ref{fig:1}(a), such a structure constrains the flexibility and robustness that are typically expected in fully distributed systems.

FL, as an emerging privacy-preserving paradigm of distributed ML, faces critical challenges such as statistical heterogeneity and privacy concerns~\cite{liFederatedLearningChallenges2020,zhuFederatedLearningNonIID2021}. Depending on communication topologies, FL can be broadly categorized into centralized and decentralized architectures~\cite{nguyenFederatedLearningInternet2021}, as illustrated in  Fig.~\ref{fig:1}. Notably, existing FL algorithms predominantly adopt centralized paradigms, with Federated Averaging (FedAvg) emerging as the de facto standard due to its simplicity and low communication overhead~\cite{mcmahanCommunicationefficientLearningDeep2017fedavg}. Nevertheless, in scenarios with significant statistical heterogeneity, where data distributions vary considerably across clients, FedAvg is prone to ``client drift" during training, resulting in unstable convergence and degraded performance~\cite{karimireddySCAFFOLDStochasticControlled2020}. To address this issue, FedProx was proposed as a representative alternative~\cite{LiTianMLSYS2020_1f5fe839}. Although FedProx claims to mitigate the adverse effects of statistical heterogeneity, our theoretical analysis and empirical evaluations under various non-independent and identically distributed (non-IID) settings reveal that it remains inadequate in stabilizing convergence and preserving performance.

Motivated by the limitations of existing approaches, where DGD-based distributed optimization algorithms typically rely on assumptions of unbiased estimation~\cite{farinaRandomizedBlockProximal2021,huangImprovingTransientTimes2023} or neglect such assumptions in their modeling, and consensus-based ADMM variants are structurally confined to two-level coordination, this study aims to develop a novel class of distributed optimization methods to address statistical heterogeneity while accommodating arbitrary communication topologies, thereby fulfilling the specific requirements of FL systems. 

The principal contributions are summarized as follows:
\begin{itemize}
    \item[a)] Algorithmic Framework: For both centralized and decentralized FL scenarios, we propose a class of distributed optimization algorithms based on the augmented Lagrangian technique. We also propose accelerated variants of the baseline algorithm, with rigorous theoretical guarantees for both standard and accelerated versions. Experimental results demonstrate that the proposed methods exhibit superior performance in large-scale distributed settings with non-IID data.
    \item[b)] Theoretical Unification: Theoretically established the downward compatibility of the proposed method, which reduces to multiple classical optimization algorithms under specific conditions, including Proximal Algorithm (PA), Gradient Descent (GD), Stochastic Gradient Descent (SGD), DGD, and consensus-based ADMM variants. Moreover, it encompasses mainstream federated optimization approaches such as FedAvg and FedProx, as well as classical unconstrained distributed optimization techniques such as Block Coordinate Descent (BCD) and its variants. This discovery bridges the theoretical gap between monolithic and distributed optimization paradigms, offering a cohesive analytical framework for cross-paradigm methodology comparisons.
\end{itemize}

The paper is structured as follows. Section~\ref{sec:2.Preliminaries} reviews related work. Section~\ref{sec:3.Fed-DALD} presents the proposed distributed optimization framework, along with its convergence analysis and accelerated variants in Section in Section~\ref{sec:4.Theoretical Analysis}. Section~\ref{sec:5.Topological Insights and Framework Unification}  provides a topological interpretation and theoretical extensions, while Section~\ref{sec:6.Numerical Experiments} details the experimental setup and results. Conclusions are drawn in Section~\ref{sec:7.Conclusion}, and relevant derivations are provided in Appendix~\ref{sec:8.Appendices} for completeness.

\section{Preliminaries}
\label{sec:2.Preliminaries}

\subsection{Related Work}
\label{subsec:Related Work}
Consider optimization problems with the following form:
\begin{equation}
\begin{aligned}
& \min_{\mathbf{x}} & & f(\mathbf{x}) \\
&~\text{s.t.} & & h(\mathbf{x}) = 0,
\end{aligned}
\label{eq:2}
\end{equation}
where \( f: \mathbb{R}^{mn} \to \mathbb{R} \) and the elements of \( h: \mathbb{R}^{mn} \to \mathbb{R}^q \) are continuous functions. Associating the Lagrange multipliers \( \mu \in \mathbb{R}^q \)  and the positive penalty vector \( \rho \in \mathbb{R}^q \) with the constraint, the Lagrangian function is defined as:
\begin{equation}
L(\mathbf{x}, \mu) = f(\mathbf{x}) + \mu^\top h(\mathbf{x})\label{eq:3}
\end{equation}
and the augmented Lagrangian has the form:
\begin{equation}
\Lambda_\rho (\mathbf{x}, \mu) = f(\mathbf{x}) + \mu^\top h(\mathbf{x}) + \| \rho \circ h(\mathbf{x}) \|^2 \label{eq:4}
\end{equation}
where the symbol \( \circ \) represents the Hadamard product in (4), i.e.,
$\mathbf{c} \circ \mathbf{d} = \left[ c_1, c_2, \dots, c_n \right]^\top \circ \left[ d_1, d_2, \dots, d_n \right]^\top = \left[ c_1 d_1, c_2 d_2, \dots, c_n d_n \right]^\top,$
and \( \| \cdot \| \) denotes the 2-norm. To tackle the problem (2), we can leverage the standard augmented Lagrangian method presented in Algorithm~\ref{alg:alm}.

\begin{algorithm}[H]
~~\caption{Augmented Lagrangian Method (ALM)} \label{alg:alm}
\begin{algorithmic}
\STATE \textbf{Initialize}: Set $k = 1$, and give the initial Lagrange multipliers $\mu^1$ and the penalty $\rho$.
\STATE \textbf{Step 1}: For fixed parameters $\mu^k$, calculate $\mathbf{x}^k$ as a solution of the problem:
\begin{equation}
    \mathbf{x}^k = \arg \min \Lambda_{\rho}(\mathbf{x}, \mu^k). \label{eq:5}
\end{equation}
\STATE \textbf{Step 2}: If $h(\mathbf{x})=0$, then stop (optimal solution found), let $\mathbf{x}^* = \mathbf{x}^k$. Otherwise, update:
\begin{equation}
    \mu^{k+1} = \mu^k + 2\rho \circ \rho \circ h(\mathbf{x}^k). \label{eq:6}
\end{equation}
\hspace{2em} Set $k = k + 1$, and repeat from Step 1.
\end{algorithmic}
\end{algorithm}

The ALM operates through two fundamental steps:
1) Inner layer (minimizing the relaxation problem): Given the Lagrange multipliers $ \mu^k $, the optimal solution $ \mathbf{x}^k $ is determined by solving $ \Lambda_\rho (\mathbf{x}, \mu) $.
2) Outer layer (updating the Lagrange multipliers): The Lagrange multipliers $ \mu^k $ are updated to $ \mu^{k+1} $. Under assumptions \hyperref[assumption:A1]{(A1)}--\hyperref[assumption:A3]{(A3)}, to be detailed later, applying the ALM to the primal problem~(\ref{eq:2}) guarantees an optimal solution~\cite{bertsekasMethodMultipliersConvex1975,bertsekasConstrainedOptimizationLagrange1996,bertsekasNonlinearProgramming2016,xuIterationComplexityInexact2021}.

\subsection{Data-Driven Problem Formulation}
Consider a ML problem involving \( N \) samples. We begin by reviewing the common formulation of such a problem, which is typically formulated as follows:
\begin{equation}
\min_{\mathbf{w}} f(\mathbf{w}) = \frac{1}{N} \sum_{i=1}^N \ell(\mathbf{w}; a_i, b_i) + \Omega(\mathbf{w}),
\label{eq:7}
\end{equation}
where \( \ell(\mathbf{w}; a_i, b_i) \) represents the loss function associated with the prediction on the sample \( (a_i, b_i) \), made using the model parameters \( \mathbf{w} \in \mathbb{R}^m \). The term \( \Omega(\mathbf{w}) \) denotes a regularization function, which may be non-smooth. This general formulation can be further formalized as problem (\ref{eq:1}), providing a foundation for the ensuing discussion.

\section{Distributed Augmented Lagrangian Decomposition for Federated Learning}\label{sec:3.Fed-DALD}
When applying the ALM to solve complex large-scale problems, the relaxed problem (\ref{eq:4}) into multiple interdependent subproblems and optimize them alternately. Specifically, it involves two main steps: 
1) The inner loop decomposes the relaxed problem (\ref{eq:4}) into multiple subproblems and solves them sequentially, yielding the optimal solution of the primal variables $\mathbf{x}^{k,*}$ given the Lagrange multipliers $\mu^k$.
2) The outer loop updates  $\mu^k$ to $\mu^{k+1}$.

This section will introduces the Distributed Augmented Lagrangian Decomposition (DALD) method for FL, which is based on the coordination of subproblems across nodes and the construction of consensus constraints. These two aspects lead to the classification of the method into two types—centralized and decentralized—corresponding to the two communication topologies illustrated in Fig.~\ref{fig:1}. 

Before delving into the details, we first define some notations. Consider a communication network consisting of $n$ client nodes, where each node holds $N_i$ user samples, with $N = \sum_{i=1}^n N_i$. The data across nodes are assumed to be independent. The overall vector  \( \mathbf{x} = (\mathbf{x}_1, \mathbf{x}_2, \dots, \mathbf{x}_n)  \in \mathbb{R}^{mn} \) represents the local parameters across all clients.\footnote{To streamline the notation, we adopt \( \mathbf{x} = (\mathbf{x}_1, \mathbf{x}_2, \dots, \mathbf{x}_n) \) in place of \( \mathbf{x} = [\mathbf{x}_1^\top, \mathbf{x}_2^\top, \dots, \mathbf{x}_n^\top]^\top \), with analogous simplifications for other vector representations.} Each subvector \( \mathbf{x}_i = (x_{i1}, x_{i2}, \dots, x_{im}) \in \mathbb{R}^m \), corresponds to the local parameters at client \( i \),  \( i \in \mathbb{N}_n  \).

\subsection{Distributed Augmented Lagrangian Decomposition with Centralized Consensus}\label{subsec:Fed-DALD-CC}

As shown in Fig.~\ref{fig:1}(a), this topology follows the traditional centralized FL paradigm, where client nodes synchronize with the central server by exchanging information. Our goal is to minimize the sum of the loss functions while ensuring consistency between the parameters at the nodes and the central server. To achieve this, we introduce the consensus constraint 
$\mathcal{C} = (\mathcal{C}_1, \mathcal{C}_2, \dots, \mathcal{C}_n) = \left[\left(\hat{\mathbf{x}} - \mathbf{x}_1\right), \left(\hat{\mathbf{x}} - \mathbf{x}_2\right) , \dots, \left(\hat{\mathbf{x}} - \mathbf{x}_n\right) \right] ,$
which establishes the coupling between nodes and facilitates coordination during distributed training. Here, \( \hat{\mathbf{x}} = (\hat{x}_1, \hat{x}_2, \dots, \hat{x}_m) \in \mathbb{R}^m \) represents the parameters to be learned at the server.

Building on this setup, we reformulate the primal problem (\ref{eq:1}) into an equivalent form as follows:
\begin{equation}
\begin{aligned}
& \min_{\mathbf{x},\hat{\mathbf{x}}}
& & \sum_{i=1}^{n} f_i (\mathbf{x}_i) \\
& \text{~s.t.}
& & \hat{\mathbf{x}} - \mathbf{x}_i = 0, ~ i \in \mathbb{N}_n,
\end{aligned}
\label{eq:8}
\end{equation}
where $f_i(\mathbf{x}_i) = \frac{1}{N} \sum_{j=1}^{N_i} \ell(\mathbf{x}_i; a_{ij}, b_{ij}) + \frac{1}{n} \Omega(\mathbf{x}_i),$
and the overall function is $f(\mathbf{x}) = \sum_{i=1}^n f_i(\mathbf{x}_i).$
 These definitions also apply similarly to the subsequent problem (\ref{eq:16}).


We introduce Lagrange multipliers \( \mu = (\mu_1, \mu_2, \dots, \mu_n) \in \mathbb{R}^{mn}, \mu_i \in \mathbb{R}^m \), associated with the consensus constraints of (\ref{eq:8}). The Lagrangian is given by
\begin{equation}
L(\mathbf{x}, \mu) = \sum_{i=1}^{n} f_i (\mathbf{x}_i) + \sum_{i=1}^{n} \mu_i^\top \mathcal{C}_i.
\label{eq:9}
\end{equation}

Next, we introduce the positive penalty  \( \rho = (\rho_1, \rho_2, \dots, \rho_n) \in \mathbb{R}^{mn}, \rho_i \in \mathbb{R}^m, i \in \mathbb{N}_n \). The augmented Lagrangian is defined as
\begin{equation}
\Lambda_\rho (\hat{\mathbf{x}}, \mathbf{x}, \mu) = \sum_{i=1}^{n} f_i (\mathbf{x}_i) + \mathcal{A}_\rho (\hat{\mathbf{x}}, \mu, \mathbf{x}),
\label{eq:10}
\end{equation}
where $\mathcal{A}_\rho (\hat{\mathbf{x}}, \mu, \mathbf{x})$
\begin{equation}
\begin{aligned}
 &= \sum_{i=1}^{n} \mathcal{A}_{\rho_i}^i (\hat{\mathbf{x}}, \mu_i, \mathbf{x}_i) \\
&= \sum_{i=1}^{n} \mu_i^\top \mathcal{C}_i + \sum_{i=1}^{n} \| \rho_i \circ \mathcal{C}_i \|^2, \\
\end{aligned}
\label{eq:11}
\end{equation}
and we define
\begin{equation}
\Lambda_{\rho_i}^i (\hat{\mathbf{x}}, \mathbf{x}_i, \mu) = f_i (\mathbf{x}_i) + \mathcal{A}_{\rho_i}^i (\hat{\mathbf{x}}, \mu_i, \mathbf{x}_i)
\label{eq:12}
\end{equation}
as the local augmented Lagrangian for client \( i \), \( i \in \mathbb{N}_n \).

We   formally introduce the first distributed optimization algorithm, termed Distributed  Augmented Lagrangian Decomposition with Centralized Consensus for FL (Fed-DALD-CC). Its framework is outlined as follows.

\begin{algorithm}[H]
\caption{~~\textbf{Fed-DALD-CC}} \label{alg:fed_dald_cc}
\begin{algorithmic}
\STATE \textbf{Initialize}: Set \( k = 1 \), \( v = 0 \), and give \( \hat{\mathbf{x}}^{1,0} \), \(\mathcal{C}^{1,0}\), \( \mu^1 \), and \( \rho \).
\textbf{Step 1.1}: Let \( v = v + 1 \), and each client \( i \) solves the local subproblem in parallel to obtain \( \mathbf{x}_i^{k,v} \):
\begin{equation}
\mathbf{x}_i^{k,v} = \arg \min_{\mathbf{x}_i} \Lambda_{\rho_i}^i (\hat{\mathbf{x}}^{k,v-1}, \mathbf{x}_i, \mu),  i \in \mathbb{N}_n. \label{eq:13}
\end{equation}
\textbf{Step 1.2}: The central server solves the consensus subproblem:
\begin{equation}
\hat{\mathbf{x}}^{k,v} = \arg \min_{\hat{\mathbf{x}}} \mathcal{A}_\rho (\hat{\mathbf{x}}, \mu^k, \mathbf{x}^{k,v}), \label{eq:14}
\end{equation}
yielding \( \hat{\mathbf{x}}^{k,v} \) and 
\( \mathcal{D}^{k,v}= \hat{\mathbf{x}}^{k,v} - \hat{\mathbf{x}}^{k,v-1} \). 
If \( \mathcal{D}^{k,v} = 0 \),  go to Step 2; otherwise, return to Step 1.1.
\STATE \textbf{Step 2}: If \( \mathcal{C}^{k,v} = 0 \), stop and set \( \mathbf{x}^* = \mathbf{x}^{k,v} \). Otherwise, each client updates:
\begin{equation}
\mu_i^{k+1} = \mu_i^k + 2 \rho_i \circ \rho_i \circ \mathcal{C}_i^{k,v},  i \in \mathbb{N}_n. \label{eq:15}
\end{equation}
\hspace{2em} 
Set \( \hat{\mathbf{x}}^{k+1,0} = \hat{\mathbf{x}}^{k,v} \), \( k = k + 1 \), \( v = 0 \), and repeat from Step 1.1.
\end{algorithmic}
\end{algorithm}

\subsection{Distributed Augmented Lagrangian Decomposition with Decentralized Consensus}\label{subsec:Fed-DALD-DC}
Unlike centralized FL, in the decentralized FL paradigm, different clients communicate and collaborate directly through a peer-to-peer (P2P) approach, without relying on any central server, as illustrated in Fig.~\ref{fig:1}(b). To establish information routing between nodes, we introduce the consensus constraint vector \( \mathcal{C} = 0 \), consisting of elements from the set $\{
\mathcal{C}_{ij} = \mathbf{x}_i - \mathbf{x}_j \mid i \in \mathbb{N}_n, j \in \mathcal{R}_i, j > i
\}$, where \( \mathcal{R}_i \) denotes the collection of all clients \( j \) that have an information routing relationship with client \( i \).  We also define the variable \( \mathbf{x}_{-i} \) to represent the coupling variable of client \( i \), and is composed of elements from the set $ \{\mathbf{x}_j \mid j \in \mathcal{R}_i\}, i \in \mathbb{N}_n.$

With these definitions in place, we   proceed to transform the primal problem (\ref{eq:1}), yielding the following equivalent form:
\begin{equation}
\begin{aligned}
& \min_{\mathbf{x}} & & \sum_{i=1}^n f_i (\mathbf{x}_i) \\
& \text{s.t.} & & \mathbf{x}_i - \mathbf{x}_j = 0,  i \in \mathbb{N}_n, j \in \mathcal{R}_i, j > i.
\end{aligned}
\label{eq:16}
\end{equation}

Then, we introduce Lagrange multipliers \( \mu =\)
$\{\mu_{ij} \in \mathbb{R}^m \mid i \in \mathbb{N}_n, j \in \mathcal{R}_i, j > i\}$ and the positive penalty  \( \rho \), consisting of elements from the set
$\{\rho_{ij} \in \mathbb{R}^m \mid i \in \mathbb{N}_n, j \in \mathcal{R}_i, j > i\}$.

The Lagrangian is given by
\begin{equation}
L(\mathbf{x}, \mu) = \sum_{i=1}^n f_i (\mathbf{x}_i) + \sum_{i=1}^n \sum_{j \in \mathcal{R}_i, j > i} \mu_{ij}^\top \mathcal{C}_{ij},
\label{eq:17}
\end{equation}

The augmented Lagrangian takes the following form
\begin{equation}
\Lambda_\rho (\mathbf{x}, \mu) = \sum_{i=1}^n f_i (\mathbf{x}_i) + \sum_{i=1}^n \sum_{j \in \mathcal{R}_i, j > i} \mathcal{A}_{\rho_{ij}}^{ij} (\mathbf{x}_i, \mu_{ij}, \mathbf{x}_j)
\label{eq:18}
\end{equation}
where $\mathcal{A}_{\rho_{ij}}^{ij} (\mathbf{x}_i, \mu_{ij}, \mathbf{x}_j)=  \mu_{ij}^{\top} \mathcal{C}_{ij} + \| \rho_{ij} \circ \mathcal{C}_{ij} \|^2.$

Furthermore, we define the local augmented Lagrangian for client $i$ as:
\begin{equation}
\begin{aligned}
  \Lambda^i_{\rho_{i}} (\mathbf{x}_i, \mathbf{x}_j, \mathbf{x}_e, \mu_i) = & \; f_i (\mathbf{x}_i) +   \sum_{j \in \mathcal{R}_i, j > i} \mathcal{A}_{\rho_{ij}}^{ij} (\mathbf{x}_i, \mu_{ij}, \mathbf{x}_j)  \\ 
   \phantom{+} + &  \sum_{e \in \mathcal{R}_i, e < i} \mathcal{A}_{\rho_{ei}}^{ei} (\mathbf{x}_e, \mu_{ei}, \mathbf{x}_i)
\end{aligned}
\label{eq:20}
\end{equation}
where $\mu_i$ is composed of elements from $\{\mu_{ij}, j \in \mathcal{R}_i, j > i\} \cup \{\mu_{ei}, e \in \mathcal{R}_i, e < i\}$, $i \in \mathbb{N}_n$. Similarly, $\rho_i$ can be derived through corresponding system interactions.

We   present the second distributed optimization algorithm, called the Distributed Augmented Lagrangian Decomposition with Decentralized Consensus for FL (Fed-DALD-DC), with its procedure outlined below.

\begin{algorithm}[H]
\caption{~~\textbf{Fed-DALD-DC}} \label{alg:fed_dald_dc}
\begin{algorithmic}
\STATE \textbf{Initialize}: Set \( k = 1 \), \( v = 0 \), and give \( \mathbf{x}^{1,0} \), \(\mathcal{C}^{1,0}\), \( \mu^1 \), and \( \rho \).
\textbf{Step 1.1}: Let \( v = v + 1 \). Given a sequence \( \mathbb{S} \), each client \( s \in \mathbb{S} \) solves its local subproblem:
\begin{equation}
\mathbf{x}_s^{k,v} = \arg \min_{\mathbf{x}_s} \Lambda^s_{\rho_{s}} (\mathbf{x}_s, \mathbf{x}_j^{k,v-1}, \mathbf{x}_e^{k,v}, \mu_s^k). \label{eq:21}
\end{equation}
\textbf{Step 1.2}: Until the last subproblem is solved, resulting in \( \mathbf{x}^{k,v} \) and 
$\mathcal{D}^{k,v} = \left\{ \mathbf{x}_i^{k,v} - \mathbf{x}_i^{k,v-1} \mid i \in \mathbb{N}_n \setminus \{1\} \right\}$.
If \( \mathcal{D}^{k,v} = 0 \), proceed to Step 2; otherwise, return to Step 1.1.
\STATE \textbf{Step 2}: If \( \mathcal{C}^{k,v} = 0 \),  stop and set \( \mathbf{x}^* = \mathbf{x}^{k,v} \). Otherwise, each client  updates:
\begin{equation}
\mu_{ij}^{k+1} = \mu_{ij}^k + 2 \rho_{ij} \circ \rho_{ij} \circ \mathcal{C}_{ij}^{k,v},  i \in \mathbb{N}_n, \, j \in \mathcal{R}_i, \, j > i. \label{eq:22}
\end{equation}
\hspace{2em} 
Set \( \mathbf{x}^{k+1,0} = \mathbf{x}^{k,v} \), \( k = k + 1 \), \( v = 0 \), and repeat from Step 1.1.
\end{algorithmic}
\end{algorithm}

\section{Theoretical Analysis: Convergence and Acceleration}
\label{sec:4.Theoretical Analysis}
In this section, we investigate the convergence properties of the Fed-DALD algorithm and its accelerated variants. Based on the topology in Fig.~\ref{fig:1} and the construction of consensus constraints in Section III, it is not difficult to conclude that Fed-DALD-CC is a special case of Fed-DALD-DC. For simplicity, we formalize problems (\ref{eq:8}) and (\ref{eq:16}) as problem (\ref{eq:2}), where the objective function is $f(\mathbf{x}) = \sum_{i=1}^{n} f_i(\mathbf{x}_i)$.

Before proceeding, we introduce some notations. Let $\{\mathbf{x}^{k,v}\}$ represent the sequence generated during the $k$-th outer loop of the Fed-DALD. The limit point or endpoint of this sequence is denoted as $\bar{\mathbf{x}}^{k,v}$, where $\bar{\mathbf{x}}_i^{k,v-1}$ refers to the element immediately preceding $\bar{\mathbf{x}}_i^{k,v}$. The sequence $\{\mathbf{x}^k\}$ is then derived from $\bar{\mathbf{x}}^{k,v}$, with $\mathbf{x}^*$ representing its limit point. Similarly, we define $\bar{\mu}^k$, $\bar{\mu}^*$, and $\{\mu^k\}$.
Given the definitions above, the subsequent analysis of the Fed-DALD method will be focused on the formulation of problem (\ref{eq:2}).


\subsection{Convergence of Fed-DALD}
\label{subsec:Convergence}
                                      
By comparing DALD with ALM, it can be observed that their primary distinction lies in the inner loop: DALD employs an alternating optimization approach to obtain the optimal solution to problem~(\ref{eq:5}). Consequently, proving the convergence of Fed-DALD reduces to demonstrating the convergence of its inner loop. Specifically, it suffices to show that, for a given value of $\mu^k$, the iterative process converges to $\mathbf{x}^{k,*}$ within the $k$-th outer loop. Once this convergence is established, the task in the outer loop is solely to update the Lagrange multipliers.

When $\mu^k$ and $\rho$ are fixed, we denote
\begin{equation}
\Lambda_{\rho} (\mathbf{x}, \mu^k) = \mathcal{F}(\mathbf{x}) = \Psi(\mathbf{x}) + \Pi(\mathbf{x}) .\label{eq:23}
\end{equation}

Consider the following unconstrained problem:
\begin{equation}
\min_{\mathbf{x} \in \mathbb{R}^{mn}}  \Psi(\mathbf{x}) + \Pi(\mathbf{x}) .\label{eq:24}
\end{equation}

\noindent \textbf{Definition 1}
\cite[Ch. 10]{rockafellar2009variational} For \( \mathcal{F}: \mathbb{R}^{mn} \to \mathbb{R} \), if the subdifferential of \( \mathcal{F} \) at a point \( \mathbf{x}^{k,*} \) satisfies \( 0 \in \partial \mathcal{F}(\mathbf{x}^{k,*}) \), then \( \mathbf{x}^{k,*} \) is  a stationary point of \( \mathcal{F} \).

We then make the following assumption:
\begin{enumerate}
\item[(A1)] 
    The function $\Psi(\mathbf{x})$ is 
    continuously 
    differentiable 
    and explicitly depends on each $\mathbf{x}_i$ for all $i \in \mathbb{N}_n$, and both $\mathcal{F}(\mathbf{x})$ and $\Pi(\mathbf{x})$ are convex functions. 
    \label{assumption:A1}
\end{enumerate}

Under Assumption \hyperref[assumption:A1]{(A1)}, the differentiability of $ \Psi(\mathbf{x}) $ and its explicit dependence on each $ \mathbf{x}_i $ ensure that the partial derivatives $ \frac{\partial \Psi}{\partial \mathbf{x}_i} $ exist for all $ i \in \mathbb{N}_n $. This, in turn, guarantees the existence of directional derivatives of $ \Psi(\mathbf{x}) $ in any direction. 

Although problem~(\ref{eq:1}) may be non-smooth, the consensus constraint enforces $\Psi(\mathbf{x})$ to retain the differentiability property required by \hyperref[assumption:A1]{(A1)}, provided the underlying network topology is connected.
Furthermore, even if problem~(\ref{eq:1}) is non-convex, the quadratic penalty term in the augmented Lagrangian may still render the resulting function convex, particularly when the penalty parameter exceeds a critical threshold~\cite{bertsekasNonlinearProgramming2016}.

\noindent \textbf{Lemma 1} Under \hyperref[assumption:A1]{(A1)}, the point $\mathbf{x}^{k,*}$ is an optimal solution of the problem $\mathcal{F}(\mathbf{x})$ over $\mathbb{R}^{mn}$ if and only if 
    \[
    -\nabla \Psi(\mathbf{x}^{k,*}) \in \partial \Pi(\mathbf{x}^{k,*}).
    \]
    
\textbf{\textit{Proof:}} See Appendix \ref{subsec:Appendix C}.

We   proceed to introduce the following two key assumptions:
\begin{enumerate}
\item[(A2)] All problems are solvable at each iteration. 
\label{assumption:A2}
\item[(A3)] The Lagrangian (\ref{eq:3}) has a saddle point $(\mathbf{x}^*, \mu^*) \in \mathbb{R}^{mn} \times \mathbb{R}^q$:
\label{assumption:A3}
\end{enumerate}
$$
L(\mathbf{x}^*, \mu) \leq L(\mathbf{x}^*, \mu^*) \leq L(\mathbf{x}, \mu^*), \forall \mathbf{x} \in \mathbb{R}^{mn}, \forall \mu \in \mathbb{R}^q.
$$

Assumption \hyperref[assumption:A2]{(A2)} ensures the existence of an optimizer that guarantees the solvability of all subproblems and 
yields a relevant solution at each iteration.
Under \hyperref[assumption:A3]{(A3)}, the strong duality relation is guaranteed, i.e., the optimal values of the primal and dual problems are equal.

\noindent \textbf{Theorem 1}
    Assume \hyperref[assumption:A1]{(A1)}–\hyperref[assumption:A2]{(A2)}. Then, every limit point of \(\{ \mathbf{x}^{k,v} \}\) minimizes \(\Lambda_\rho(\mathbf{x}, \mu^k)\) over \(\mathbb{R}^{mn}\).

\textbf{\textit{Proof:}}
Denote 
$$
  \varpi_i^{k,v} = (\mathbf{x}_1^{k,v+1}, \dots, \mathbf{x}_{i-1}^{k,v+1}, \mathbf{x}_i^{k,v+1}, \mathbf{x}_{i+1}^{k,v}, \dots, \mathbf{x}_n^{k,v}),  i \in \mathbb{N}_n.
$$
According to \hyperref[assumption:A2]{(A2)}, given the current iterate \( \mathbf{x}^{k,v} = (\mathbf{x}_1^{k,v}, \mathbf{x}_2^{k,v}, \dots, \mathbf{x}_n^{k,v}) \), we compute the next iterate \( \mathbf{x}^{k,v+1} = (\mathbf{x}_1^{k,v+1}, \mathbf{x}_2^{k,v+1}, \dots, \mathbf{x}_n^{k,v+1}) \) via DALD.  Based on the iterative minimization of the subproblems, i.e., (\ref{eq:13}), (\ref{eq:14}), or (\ref{eq:21}), we derive
\begin{equation}
\fontsize{9.5}{12}\selectfont
\begin{aligned}
  \mathcal{F}(\mathbf{x}^{k,v}) \geq \mathcal{F}(\varpi_1^{k,v}) \geq \mathcal{F}(\varpi_2^{k,v}) \geq \dots \geq \mathcal{F}(\varpi_{n-1}^{k,v})  
  \geq \mathcal{F}(\mathbf{x}^{k,v+1}) \label{eq:25}
\end{aligned}
\end{equation}
Let \( \bar{\mathbf{x}}^{k,v} = (\bar{\mathbf{x}}_1^{k,v}, \bar{\mathbf{x}}_2^{k,v}, \dots, \bar{\mathbf{x}}_n^{k,v}) \) be a limit point of the sequence \( \{ \bar{\mathbf{x}}^{k,v} \} \). Equation (\ref{eq:25}) implies that the sequence \( \{ \mathcal{F}(\mathbf{x}^{k,v}) \} \) converges to \( \mathcal{F}(\bar{\mathbf{x}}^{k,v}) \). We will   demonstrate that \( \bar{\mathbf{x}}^k \) satisfies the optimality condition:
\[
\Psi(\bar{\mathbf{x}}^{k,v}) + \Pi(\bar{\mathbf{x}}^{k,v}) \leq \Psi(\mathbf{x}) + \Pi(\mathbf{x}), \forall \mathbf{x} \in \mathbb{R}^{mn}.
\]
According to Lemma 1, we know that it suffices to prove that
\[
-\nabla \Psi(\bar{\mathbf{x}}^{k,v}) \in \partial \Pi(\bar{\mathbf{x}}^{k,v}).
\]
Consider the sequence \( \{ \mathbf{x}^{k,v} \} \) converging to \( \mathbf{x}^{k,v} \). According to DALD algorithm and~\eqref{eq:25}, we deduce:
$$
  \mathcal{F}(\mathbf{x}^{k,v+1}) \leq \mathcal{F}(\varpi_1^{k,v}) \leq \mathcal{F}(\mathbf{x}_1, \mathbf{x}_2^{k,v}, \dots, \mathbf{x}_n^{k,v}),  \forall \mathbf{x}_1 \in \mathbb{R}^m.
$$
Taking the limit as \( v \to \infty \), we get
\[
\mathcal{F}(\bar{\mathbf{x}}^{k,v}) \leq \mathcal{F}(\mathbf{x}_1, \bar{\mathbf{x}}_2^{k,v}, \dots, \bar{\mathbf{x}}_n^{k,v}),  \forall \mathbf{x}_1 \in \mathbb{R}^m.
\]
Consequently, for the function \( \mathcal{F}(\mathbf{x}_1, \bar{\mathbf{x}}_2^{k,v}, \dots, \bar{\mathbf{x}}_n^{k,v}) \) with respect to \( \mathbf{x}_1 \), there exists an optimum \( \mathbf{x}_1^{k,v} \). According to~\hyperref[assumption:A1]{(A1)} and~\hyperref[lemma:1]{Lemma 1}, we obtain:
\[
-\nabla_1 \Psi(\bar{\mathbf{x}}^{k,v}) \in \partial_1 \Pi(\bar{\mathbf{x}}^{k,v}),
\]
where \( \nabla_i \) denotes the gradient of \( \Psi \) and \( \partial_i \) denotes the subdifferential of \( \Pi \) with respect to \( \mathbf{x}_i \).
Similarly, for each component \( \mathbf{x}_i \):
\[
-\nabla_i \Psi(\bar{\mathbf{x}}^{k,v}) \in \partial_i \Pi(\bar{\mathbf{x}}^{k,v}),  i \in \mathbb{N}_n. 
\]
By further consolidating:
\begin{equation}
\begin{aligned}
  -\left[\nabla_1 \Psi(\bar{\mathbf{x}}^{k,v}), \nabla_2 \Psi(\bar{\mathbf{x}}^{k,v}), \dots, \nabla_n \Psi(\bar{\mathbf{x}}^{k,v})\right] 
  \in \\
  \left\{(g_1, g_2, \dots, g_n)\mid g_i \in \partial_i \Pi(\bar{\mathbf{x}}^{k,v}), \ i \in \mathbb{N}_n \right\},
\end{aligned}
\label{eq:26}
\end{equation}
which can be rewritten as:
\[
-\nabla \Psi(\bar{\mathbf{x}}^{k,v}) \in \partial \Pi(\bar{\mathbf{x}}^{k,v}).
\]
This concludes the proof. \hfill $\square$

\textit{Remark 1:}
From the proof of~\hyperref[theorem:1]{Theorem 1}, it is evident that during the iterative process of the algorithm, each subproblem must be traversed continuously to ensure convergence to the optimal point. Therefore, In the distributed optimization framework (DALD), consensus among all nodes requires that the communication topology be connected.

\textit{Remark 2:}
Consider two functions \( f: \mathcal{X} \to \mathbb{R} \) and \( \mathcal{F}: \mathcal{X} \to \mathbb{R} \), where the surrogate function \( \mathcal{F} \), constructed (e.g., using (\ref{eq:18})), exhibits better properties than \( f \) (e.g., \( \mathcal{F} \) satisfies condition~\hyperref[assumption:A1]{(A1)}, whereas \( f \) does not). Since \( f \) and \( \mathcal{F} \) share the same minimum points,
alternating optimization of \( \mathcal{F} \) is, based on~\hyperref[theorem:1]{Theorem 1}, equivalent to minimizing \( f \). This simple yet fundamental idea is at the core of this paper.

\textit{Remark 3:}
    When the convexity assumption of $\mathcal{F}$ in~\hyperref[assumption:A1]{(A1)} is not satisfied,~\hyperref[theorem:1]{Theorem 1} can only guarantee stationary point, not necessarily global optima.

\subsection{Acceleration of Fed-DALD}
When configuring the stopping criteria for the inner and outer loops, distinct tolerances are typically assigned: $\epsilon_{\text{dual}}$ for the inner loop and $\epsilon_{\text{pri}}$ for the outer loop. These tolerances represent predefined thresholds for computational accuracy. Concretely, the inner loop stops when the condition 
\begin{equation*}
    \text{(B1)} ~~\|\mathcal{D}^{k,v} \|_\infty \leq \epsilon_{\text{dual}} \approx 0
    \label{condition:B1}
\end{equation*}
is satisfied, ensuring that the outer loop receives a solution from the inner loop with a sufficiently accurate level. However, empirical evidence suggests that setting an excessively high precision at the initial stages leads to unnecessary computational resource consumption. 

As noted by Bertsekas~\cite{bertsekasCombinedPrimaldualPenalty1975}, 
the problem (\ref{eq:5}) in the inner loop is typically required to be solved exactly by default.
Yet, even if the minimization process terminates prematurely, the algorithm may still converge. This observation prompted us to reconsider the necessity of obtaining an exact solution $\mathbf{x}^{k,*}$ in the initial phase of DALD. Hence, a modified stopping criterion is integrated into Step 1.2 of the DALD algorithm to improve computational efficiency:
\begin{enumerate}
\label{condition:B2}
\item[(B2)] If  $\|\mathcal{D}^{k,v} \|_\infty \leq \epsilon_{\text{dual}}^k$ holds, where $\{\epsilon_{\text{dual}}^k\}$ satisfies $0 \leq \epsilon_{\text{dual}}^k, \forall k$, and $\epsilon_{\text{dual}}^k \rightarrow \epsilon_{\text{dual}}$, proceed to Step 2.
\end{enumerate}

Next, we will prove the convergence of the proposed DALD variant, with particular focus on the stopping criterion for the inner loop (i.e., condition (B2)). Based on Appendix A, we define
$h(\mathbf{x}^k) = h(\bar{\mathbf{x}}^{k,v})=\mathcal{C}^k $
and
$\mathcal{D}^k = \left\{ \mathcal{D}_i^k = \bar{\mathbf{x}}_i^{k,v} - \bar{\mathbf{x}}_i^{k,v-1} \mid i \in \mathbb{N}_n \setminus \{1\} \right\}.$
These definitions capture the residual information of the solution at the end of the \(k\)-th outer loop. The same definitions extend to the final solution at the end of the outer loop iteration, i.e., the limit point \(\mathbf{x}^*\) of the sequence \(\{\mathbf{x}^k\}\), yielding a similar definition for \(h(\mathbf{x}^*)\).

\noindent \textbf{Theorem 2} 
Assume (A1)–(A3). For \(k = 0, 1, \dots\), let \(\{\mathbf{x}^{k,v}\}\) satisfy
\[
\|\mathcal{D}^{k,v}\|_\infty \leq \epsilon_{\text{dual}}^k,
\]
and suppose that \(\{\mu^k\}\) is bounded, \(\{\epsilon^k\}\) and \(\{\rho^k\}\) satisfy
\[
0 < \rho^k < \rho^{k+1}, \quad \forall k, \quad \rho^k \to \infty,
\]
\[
0 \leq \epsilon_{\text{dual}}^k, \quad \forall k, \quad \epsilon_{\text{dual}}^k \to 0.
\]
Then,
\[
\{\mu^k + 2\rho^k \circ \rho^k \circ h(\mathbf{x}^k)\}_K \to \mu^*,
\]
where \(\mu^*\) is a vector satisfying, together with \(\mathbf{x}^*\), the following conditions:
\[
- \nabla h(\mathbf{x}^*)^\top \mu^* \in \partial f(\mathbf{x}^*), \quad h(\mathbf{x}^*) = 0.
\]
Thus, \(\mathbf{x}^*\) is a minimizer.

\begin{figure*}[t!]
    \centering
    \includegraphics[width=0.95\linewidth, trim=0.5cm 0.1cm 0.5cm 0cm, clip]{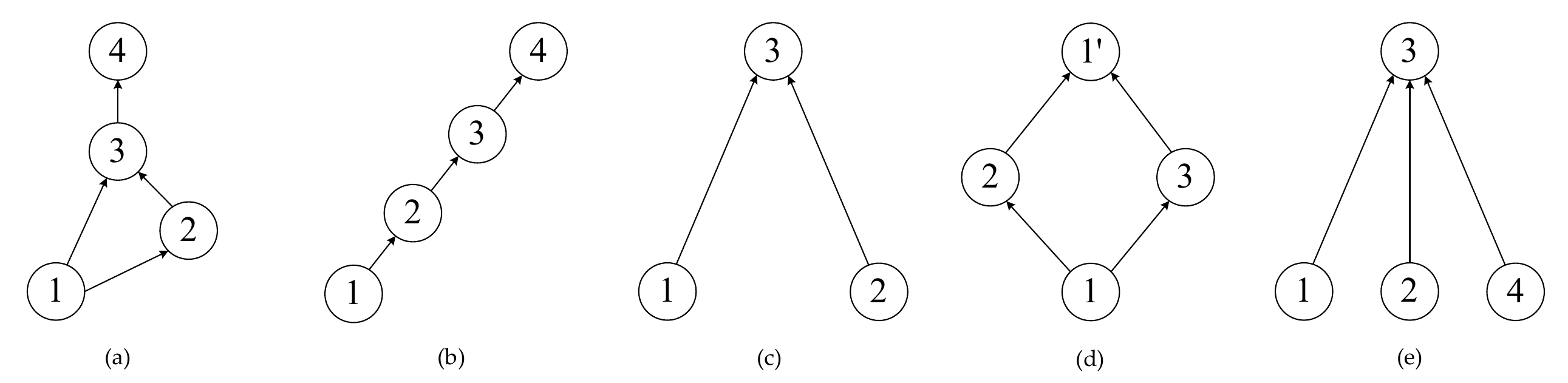}
        \caption{Hierarchical Networks for Client Coordination Sequences}
    \label{fig:2}
\end{figure*}

\begin{figure*}[t!]
    \centering
    \includegraphics[width=0.95\linewidth, trim=0.5cm 0.1cm 0.5cm 0cm]{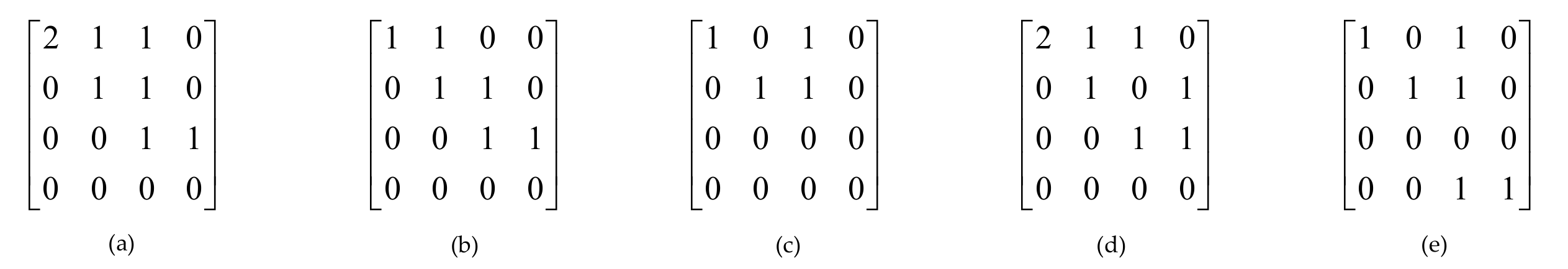}
        \caption{Hierarchical Matrices for Client Coordination Sequences}
    \label{fig:3}
\end{figure*}

\textbf{\textit{Proof:}}
Without loss of generality, assume that the entire sequence \(\{\mathbf{x}^k\}\) converges to \(\mathbf{x}^*\). Define
\[
\mu^{k+1} = \mu^k + 2\rho^k \circ \rho^k \circ h(\mathbf{x}^k).
\]
From this definition, the subdifferential of the augmented Lagrangian function (\ref{eq:3}) is
\begin{equation}
    \partial \Lambda_{\rho^k} (\mathbf{x}^k, \mu^k) = \partial f(\mathbf{x}^k) + \nabla h(\mathbf{x}^k)^\top \mu^{k+1}. 
     \label{eq:27}
\end{equation}
Rearranging terms gives
\[
\nabla h(\mathbf{x}^k)^\top \mu^{k+1} = \partial \Lambda_{\rho^k} (\mathbf{x}^k, \mu^k) - \partial f(\mathbf{x}^k).
\]
Multiplying both sides by \(\left[ \nabla h(\mathbf{x}^k) \nabla h(\mathbf{x}^k)^\top \right]^{-1} \nabla h(\mathbf{x}^k)\), we obtain
\begin{equation}
\begin{aligned}
    \mu^{k+1} =  &  \left[ \nabla h(\mathbf{x}^k)   \nabla h(\mathbf{x}^k)^\top \right]^{-1} \nabla h(\mathbf{x}^k) \\
     & \quad \quad \left[ \partial \Lambda_{\rho^k} (\mathbf{x}^k, \mu^k) - \partial f(\mathbf{x}^k) \right]. 
     \label{eq:28}
\end{aligned}
\end{equation}
Following the properties of convex functions and Theorem  1, as \(k \to \infty\) with \(\epsilon_{\text{dual}}^k \to \epsilon_{\text{dual}} \approx 0\), there exists a limit point \(\mathbf{x}^*\) such that \(g_i^k \in \partial_i \Lambda_{\rho^k} (\mathbf{x}^k, \mu^k)\) with
\begin{equation}
   g_i^k \to 0, i\in \mathbb{N}_n. \label{eq:29}
\end{equation}

Furthermore, from (\ref{eq:28}), we have
\[
\mu^{k+1} \to \mu^*, 
\]
where
\[
\mu^* = -\left[ \nabla h(\mathbf{x}^*) \nabla h(\mathbf{x}^*)^\top \right]^{-1} \nabla h(\mathbf{x}^*) \partial f(\mathbf{x}^*).
\]
Given that \(g^k \to 0\) and by (\ref{eq:27}), it follows that
\[
0 \in \partial f(\mathbf{x}^*) + \nabla h(\mathbf{x}^*)^\top \mu^*.
\]
Since \(\{\mu^k\}\) is bounded and \(\mu^k + 2\rho^k \circ \rho^k \circ h(\mathbf{x}^k) \to \mu^*\), it follows that \(\{2\rho^k \circ \rho^k \circ h(\mathbf{x}^k)\}\) remains bounded. As \(\rho^k \to \infty\), we deduce \(h(\mathbf{x}^k) \to 0\), which implies \(h(\mathbf{x}^*) = 0\), confirming that \(\mathbf{x}^*\) is an optimal solution. \hfill $\square$

Next, we present two termination criteria: the first aligns with condition~\hyperref[condition:B2]{(B2)}, while the second offers a more relaxed alternative. These criteria are designed to facilitate the practical implementation of our algorithm. 
\begin{itemize}
    \label{condition:B3}
    \item[(B3)] If $v = v_{\max}^k$ or $\|\mathcal{D}^{k,v}\|_\infty \leq \epsilon_{\text{dual}}$ holds, where $\{v_{\max}^k\}$ satisfies $1 \leq v_{\max}^k$, $\forall k$, $v_{\max}^k \to \infty$, proceed to Step 2.
\end{itemize}
\begin{itemize}
    \label{condition:B4}
    \item[(B4)] If $v = v_{\max}$ or $\|\mathcal{D}^{k,v}\|_\infty \leq \epsilon_{\text{dual}}$ holds, proceed to Step 2.
\end{itemize}

When establishing the termination conditions~\hyperref[condition:B2]{(B2)}, \hyperref[condition:B3]{(B3)}, and \hyperref[condition:B4]{(B4)}, it is crucial to revise the stopping criterion for the outer loop as follows:
\begin{equation*}
     \text{If } \|\mathcal{C}^{k,v}\|_\infty \leq \epsilon_{\text{pri}} \text{ and } \|\mathcal{D}^{k,v}\|_\infty \leq \epsilon_{\text{dual}} \text{ hold}\dots
\end{equation*}
This adjustment ensures that the algorithm does not terminate prematurely before $\mathcal{D}^{k,v}$ fulfills the required condition.

\section{Topological Insights and Framework Unification}
\label{sec:5.Topological Insights and Framework Unification}
\subsection{Topological Interpretation}
As shown in Fig.~\ref{fig:1}(a), Centralized FL adopts a fixed communication topology, where all clients optimize their local parameters in parallel and then transmit the information to the central server to reach consensus. However, when communication conditions (e.g., connectivity and resource availability) permit, any node can dynamically establish information routing paths, enabling clients in Decentralized FL to communicate and collaborate directly through P2P connections without relying on a central server, as illustrated in Fig.~\ref{fig:1}(b). Based on such networks, we define the client coordination sequence $\mathbb{S}$ and describe its structure using Hierarchical Network and Hierarchical Matrix, two concepts initially proposed in our prior work~\cite{guoDistributedAugmentedLagrangian2025}.

\noindent \textbf{Definition 2}
    A \textit{hierarchical network} is built from a root node, recursively branching to lower-level nodes. Each node is assigned a level and a unique identifier until all nodes are integrated.

\noindent \textbf{Definition 3}
    A \textit{hierarchical matrix} \( \mathcal{H} = [a_{ij}] \) is an \( n \times n \) matrix encoding hierarchical relationships. If node \( i \) is a direct descendant of node \( j \), \( a_{ij} = 1 \); otherwise, \( a_{ij} = 0 \). The diagonal \( a_{ii} \) represents the out-degree of node \( i \).

In the hierarchical network, directed edges represent the flow of information, passing from lower-level clients to higher-level clients. A parent client can only start its computation after all its dependent lower-level clients have been completed and their information collected. The subproblem at the root node is the last one to be solved, while the nodes in each subsequent level represent subproblems that need to be solved before those in higher levels. 
Hence, $j > i$ in (\ref{eq:16}) indicates that subproblem $j$ resides at a higher hierarchical level than subproblem $i$, and the two are interconnected within the hierarchical network.
Fig.~\ref{fig:2} shows some client coordination sequences represented by hierarchical networks, and Fig.~\ref{fig:3} provides a detailed description of the corresponding sequences using hierarchical matrices.

In the implementation of Fed-DALD, \textit{full-cycle coordination }is used by default, meaning that all subproblems are handled in each inner loop, as shown in Fig.~\ref{fig:2}(a), (b), and (e). To meet finer control requirements, the algorithm also supports \textit{partial-cycle coordination}, where a portion or a single subproblem is processed in each iteration, as shown in Fig.~\ref{fig:2}(c), and \textit{selective-repetitive coordination}, which involves revisiting certain subproblems multiple times, as shown in Fig.~\ref{fig:2}(d). Essentially, the latter can be achieved by combining the first two coordination types.

The selection of subproblems can be based on random or greedy strategies, but according to ~\hyperref[remark:1]{Remark 1}, each subproblem should be considered with equal probability for potential optimization, although the actual selection may vary depending on the strategy. This prevents over-sampling or under-sampling, ensuring correct convergence.

Moreover, Fed-DALD supports online configuration of client coordination sequences in different inner loops, enabling strong fault tolerance for scenarios like network instability, client computing fluctuations, data quality issues, and unbalanced participation. For example, if clients drop out or fail to transmit information, the system dynamically adjusts the hierarchical coordination network to mitigate dropout effects. 
To ensure smooth termination and consistent dual variable updates, it is recommended that the last inner loop of the current outer loop employ full-cycle coordination, aligning with the initial sequence \(\mathbb{S}^{k,1}\), particularly when utilizing partial-cycle or selective-repetitive coordination strategies.

\subsection{Framework Unification}

\label{subec:5.2Framework Unification}

Motivated by Remark 2, this section focuses on the Fed-DALD-CC framework, where we propose two alternative strategies, aside from augmented Lagrangian relaxation: (a) incorporation of proximal regularization and (b) construction of a second-order approximation. These approaches are employed to construct a surrogate function for $f(\mathbf{x})$, which is then alternately optimized to solve the original problem. During this process, we systematically unify and derive existing unconstrained optimization methods. Intriguingly, canonical techniques (e.g., PA and GD) emerge as special cases of our proposed methodology, thereby establishing a unified perspective for both classical monolithic optimization and distributed optimization.

To streamline the analysis, we assume uniform penalty parameters: \(\alpha = \frac{1}{2\rho_{ir}^2}\), \(i \in \mathbb{N}_n\), \(r \in \mathbb{N}_m\). The closed-form solution to subproblem (\ref{eq:14}) is derived as:
\begin{equation}
    \hat{\mathbf{x}}^{k,v} = \frac{\sum \rho_i \circ \rho_i \circ \mathbf{x}_i^{k,v} - \frac{1}{2} \sum \mu_i^k}{\sum \rho_i \circ \rho_i} = \frac{\sum \mathbf{x}_i^{k,v} - \alpha \sum \mu_i^k}{n}.
    \label{eq:31}
\end{equation}

\subsubsection{Monolithic Optimization}

Initially, let $\mu = 0$ and $n = 1$. The augmented Lagrangian (\ref{eq:10}) can be rewritten as
\begin{equation}
\mathcal{F} = f(\mathbf{x}) + \frac{1}{2\alpha} \|\hat{\mathbf{x}} - \mathbf{x}\|^2.
\label{eq:32}
\end{equation}


Reformulate (\ref{eq:13}) and (\ref{eq:31}) as:
\begin{equation}
\mathbf{x}^k = \arg \min_{\mathbf{x}} \left\{ f(\mathbf{x}) + \frac{1}{2\alpha} \|\hat{\mathbf{x}}^{k-1} - \mathbf{x}\|^2 \right\},
\label{eq:33}
\end{equation}
\begin{equation}
\hat{\mathbf{x}}^k = \mathbf{x}^k.
\label{eq:34}
\end{equation}

\begin{figure*}[t]
    \centering
    \includegraphics[width=0.98\linewidth, trim=0cm 0cm 0cm 0cm]{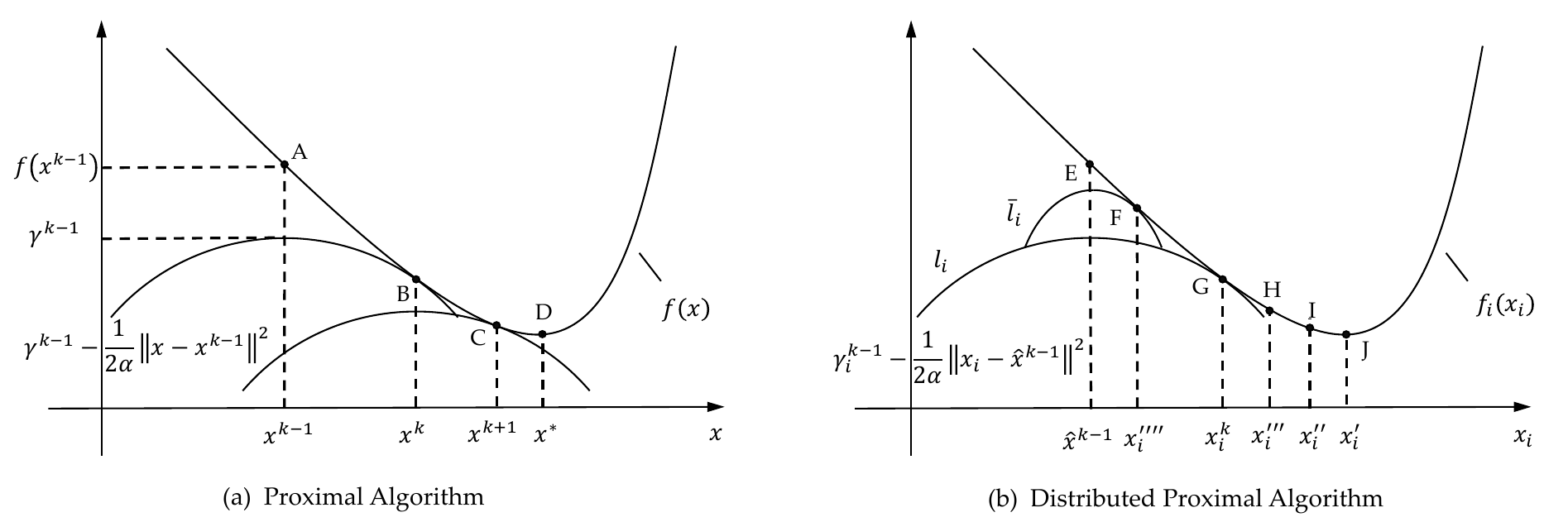}
    \caption{Geometric View of the Proximal Framework}
    \label{fig:4}
\end{figure*}

Substituting (\ref{eq:34}) into (\ref{eq:33}), we obtain
\begin{equation}
\mathbf{x}^{k+1} = \arg \min_{\mathbf{x}} \left\{ f(\mathbf{x}) + \frac{1}{2\alpha} \|\mathbf{x} - \mathbf{x}^k\|^2 \right\}.
\end{equation}
This leads to the well-known PA~\cite{bertsekasConvexOptimizationAlgorithms2015, bertsekasNonlinearProgramming2016}, with the iteration mechanism illustrated in Fig.~\ref{fig:4}(a). 
Combining this with (\ref{eq:A1-3}), it becomes evident that PA is a special case of Fed-DALD-CC when $n=1$ and $\mu = \mu^* = 0$.

Next, we suppose that \( f \) is convex and twice continuously differentiable. Set \(\mu = \nabla f(\mathbf{x})\) and \(n = 1\), and we define the surrogate function \(\mathcal{F}\) as:
\begin{equation}
\mathcal{F} 
= f(\mathbf{x}) + \nabla f(\mathbf{x})^{\top} (\hat{\mathbf{x}} - \mathbf{x}) + \frac{1}{2\alpha} \|\hat{\mathbf{x}} - \mathbf{x}\|^2.
\label{eq:36}
\end{equation} 

Let \(\mathbf{x}^k\) minimize \(\mathcal{F} (\hat{\mathbf{x}}^{k-1}, \mathbf{x}, \mu^k)\). This results in
\begin{equation}
\left[ \nabla^2 f(\mathbf{x}^k) - \frac{I}{\alpha} \right] (\mathbf{x}^k - \hat{\mathbf{x}}^{k-1}) = 0.
\label{eq:37}
\end{equation}
It is straightforward to deduce that:
\begin{equation}
\mathbf{x}^k = \hat{\mathbf{x}}^{k-1}.
\label{eq:38}
\end{equation}

Let \(\hat{\mathbf{x}}^k\) minimize \(\mathcal{F} (\hat{\mathbf{x}}, \mathbf{x}^k, \mu^k)\).
We then obtain:
\begin{equation}
\hat{\mathbf{x}}^k = \mathbf{x}^k - \alpha \nabla f(\mathbf{x}^k).
\label{eq:39}
\end{equation}
According to (\ref{eq:38}), it follows that:
\begin{equation}
\mathbf{x}^{k+1} = \mathbf{x}^k - \alpha \nabla f(\mathbf{x}^k).
\label{eq:40}
\end{equation}
This is the well-known GD Algorithm.

Next, from (\ref{eq:37}), if \(\nabla^2 f(\mathbf{x}^k)\) is nonsingular, we deduce that:
\begin{equation}
\alpha^k I = \nabla^2 f(\mathbf{x}^k)^{-1}.
\label{eq:41}
\end{equation}

Using the fact (\ref{eq:36}) and (\ref{eq:40}), and performing a second-order approximation with the latest information \( \mathbf{x}^k \), we obtain:
\begin{equation}
\mathbf{x}^{k+1} = \mathbf{x}^k - \nabla^2 f(\mathbf{x}^k)^{-1} \nabla f(\mathbf{x}^k).
\label{eq:42}
\end{equation}
This is the familiar Newton's Method (NM).

By Theorem 1, we assume that (\ref{eq:40}) and (\ref{eq:42}) convergence to the point \(( \bar{\mathbf{x}}^k, \bar{\hat{\mathbf{x}}}^k )\). Consequently, the gradient conditions of \(\mathcal{F}\) yield:
\begin{equation}
\nabla_{\mathbf{x}} \mathcal{F} =  \nabla^2 f(\bar{\mathbf{x}}^k)(\bar{\hat{\mathbf{x}}}^k - \bar{\mathbf{x}}^k)  - \frac{1}{\alpha} (\bar{\hat{\mathbf{x}}}^k - \bar{\mathbf{x}}^k) = 0,
\label{eq:43}
\end{equation}
\begin{equation}
\nabla_{\hat{\mathbf{x}}} \mathcal{F} = \nabla f(\bar{\mathbf{x}}^k) + \frac{1}{\alpha} (\bar{\hat{\mathbf{x}}}^k - \bar{\mathbf{x}}^k) = 0.
\label{eq:44}
\end{equation}
From (\ref{eq:43}), we derive:
\begin{equation*}
\left[ \nabla^2 f(\bar{\mathbf{x}}^k) - \frac{I}{\alpha} \right] (\bar{\hat{\mathbf{x}}}^k - \bar{\mathbf{x}}^k) = 0.
\end{equation*}
Regardless of whether \(\nabla^2 f(\bar{\mathbf{x}}^k) - \frac{I}{\alpha}\) is zero, by employing (\ref{eq:38}) for iteration, we obtain:
\begin{equation}
\bar{\hat{\mathbf{x}}}^k - \bar{\mathbf{x}}^k = \bar{\hat{\mathbf{x}}}^k - \bar{\hat{\mathbf{x}}}^{k-1} = 0.
\label{eq:45}
\end{equation}
Substituting into (\ref{eq:44}) yields:
\begin{equation}
\nabla f(\bar{\mathbf{x}}^k) = 0.
\label{eq:46}
\end{equation}
Given the convexity of \( f \), it follows that \(\bar{\mathbf{x}}^k\) is a global minimum point of \( f \). Therefore, the convergence of both GD and NM is ensured. For GD, the fixed step size $\alpha$ needs to satisfy appropriate selection conditions, as established in~\cite{bertsekasNonlinearProgramming2016} and~\cite{ruszczynskiNonlinearOptimization2006}. Moreover, according to Theorem 2, it is theoretically expected that \(\alpha^k \to 0^+\) as \(k \to \infty\).

\subsubsection{Stochastic Optimization and Distributed Optimization for ML}

Now, we assume \( \mu = \nabla f(\mathbf{x}) \) and \( n \geq 2 \), and \( f \) is a twice continuously differentiable convex function. The surrogate function \( \mathcal{F} \) is defined as:
\begin{equation}
\begin{aligned}
        \mathcal{F} = \sum_{i=1}^n f_i(\mathbf{x}_i) +    \sum_{i=1}^n \nabla f_i(\mathbf{x}_i)^{\top} (\hat{\mathbf{x}} - \mathbf{x}_i)  
         + \frac{1}{2} \sum_{i=1}^n \frac{\|\hat{\mathbf{x}} - \mathbf{x}_i\|^2}{\alpha_i}.
\end{aligned}
\label{eq:47}
\end{equation}

Let \( \mathbf{x}_i^k \) minimize \( \mathcal{F}(\hat{\mathbf{x}}^{k-1}, \mathbf{x}, \mu^k),  i \in \mathbb{N}_n \). This yields:
\begin{equation}
\left[ \nabla_i^2 f_i(\mathbf{x}_i^k) - \frac{I}{\alpha_i} \right] (\mathbf{x}_i^k - \hat{\mathbf{x}}^{k-1}) = 0.
\label{eq:48}
\end{equation}
It is straightforward to deduce that:
\begin{equation}
\mathbf{x}_i^k = \hat{\mathbf{x}}^{k-1}.
\label{eq:49}
\end{equation}

Let \( \hat{\mathbf{x}}^k \) minimize \( \mathcal{F} (\hat{\mathbf{x}}, \mathbf{x}^k, \mu^k) \). This leads to:
\begin{equation}
\hat{\mathbf{x}}^k = \mathbf{x}_i^k - \alpha_i \nabla f_i(\mathbf{x}_i^k).
\label{eq:50}
\end{equation}
Substituting (\ref{eq:49}) into (\ref{eq:50}), we further deduce:
\begin{equation}
\hat{\mathbf{x}}^k = \hat{\mathbf{x}}^{k-1} - \alpha_i \nabla f_i(\mathbf{x}_i^k).
\label{eq:51}
\end{equation}
Alternatively, this can be expressed as:
\begin{equation}
\mathbf{x}_i^{k+1} = \mathbf{x}_i^k - \alpha_i \nabla f_i(\mathbf{x}_i^k). 
\label{eq:52}
\end{equation}

Suppose we divide the \( N \) samples into \( n \) batches or clients, where each batch contains \( N_i \) samples. Under the Fed-DALD-CC framework, implementing partial-cycle coordination, if we randomly select only one batch or client for update in the  solving step (\ref{eq:48}), then algorithm (\ref{eq:51}) reduces to Mini-batch Gradient Descent (MBGD). Further, if we treat each batch or client as an independent system, and each sample as an individual client, with only one sample selected for optimization at each step (i.e., \( B = 1 \)), or when \( N_i = 1 \), (\ref{eq:51}) further reduces to SGD. Similarly, by replacing the step size with \( \alpha_i^k I = \nabla_i^2 f(\mathbf{x}_i^k )^{-1} \), we obtain the Stochastic Newton’s Method (SNM). Analogous to (\ref{eq:43})--(\ref{eq:46}), it is straightforward to show that the above methods is convergent.

Based on the above analysis, we argue that methods such as MBGD and SGD, which 
rely on stochastic selection of data samples, 
can be considered a class of distributed optimization methods. Since the objective function \( f \) possesses favorable properties, gradient information can be directly utilized in the solver layer for efficient updates.
However, equation (\ref{eq:51}) explicitly shows that these methods also have a clear limitation, namely that they cannot efficiently utilize information from multiple clients simultaneously during each iteration.

To address the aforementioned dilemma, we  first return to the proximal framework. We set \( \mu_i = 0, i \in \mathbb{N}_n \) and \( n \geq 2 \). The step-size parameters are defined as \( \alpha_i = \frac{1}{2\rho_{ir}^2} = \frac{1}{2\beta_i} \), where \( \beta_i > 0 \), \( i \in \mathbb{N}_n, r \in \mathbb{N}_m \). Under these settings, the surrogate function \( \mathcal{F} \) is obtained as:
\begin{equation}
\mathcal{F} = \sum_{i=1}^n f_i(\mathbf{x}_i) + \frac{1}{2} \sum_{i=1}^n \frac{\|\hat{\mathbf{x}} - \mathbf{x}_i\|^2}{\alpha_i}.
\label{eq:53}
\end{equation}
Define the weights as \( p_i^k = \frac{\beta_i^k}{\sum_{i=1}^n \beta_i^k},i \in \mathbb{N}_n \). Reformulating (\ref{eq:13}) and (\ref{eq:14}), we get:
\begin{equation}
\mathbf{x}_i^k = \arg\min_{\mathbf{x}_i} \left\{ f_i(\mathbf{x}_i) + \frac{1}{2\alpha_i} \|\hat{\mathbf{x}}^{k-1} - \mathbf{x}_i\|^2 \right\},  i \in \mathbb{N}_n.
\label{eq:54}
\end{equation}
\begin{equation}
\hat{\mathbf{x}}^k = \sum_{i=1}^n p_i^k \mathbf{x}_i^k.
\label{eq:55}
\end{equation}
The algorithm defined by (\ref{eq:54}) and (\ref{eq:55}) is referred to as the Distributed PA in this paper. Notably, when \( \alpha_i \) is set as a uniform constant, this framework lead to the well-known FL algorithm, FedProx~\cite{LiTianMLSYS2020_1f5fe839}, with its iterative mechanism illustrated in Fig.~\ref{fig:4}(b). On this basis, if \( f_i \) is continuously differentiable, 
and the penalty  \( \rho_{ir} \to 0^+, i \in \mathbb{N}_n, r \in \mathbb{N}_m \), 
the FedProx further reduces to FedAvg~\cite{mcmahanCommunicationefficientLearningDeep2017fedavg}. Comparing the two and recalling Theorem 1, it can be observed that the proximal term introduced in FedProx, as compared to FedAvg, better satisfies (A1).

Under the assumption that the data across all clients satisfies the IID condition, and that each client has a sufficiently large dataset and a complete and well-designed training process, theoretically, a single client’s data is sufficient to train the global optimal parameters, allowing the local model to fully represent the global model. Specifically, as shown in Fig.~\ref{fig:4}(b), 
for all clients, their loss functions $f_i$ have identical graphs,
and the optimal solution will converge to the same point J. If $f_i$ is differentiable, then $\nabla f_i (\mathbf{x}_i^*) = 0$. 
In conjunction with PA, solving (\ref{eq:54}) is equivalent to minimizing $f_i (\mathbf{x}_i)$.
To minimize $f_i (\mathbf{x}_i)$, parameter optimization can be achieved through the iterative update formula (\ref{eq:40}):
\begin{equation}
\mathbf{x}_i^{k+1} = \mathbf{x}_i^{k} - \alpha_i^k \nabla f_i (\mathbf{x}_i^k),  i \in \mathbb{N}_n, \label{eq:56}
\end{equation}
which aligns with (\ref{eq:52}), i.e., MBGD. Without loss of generality, when $n$ clients participate in joint training, the update rule, in conjunction with (\ref{eq:55}),  can be expressed as:
\begin{equation}
\mathbf{x}_i^{k+1} = \sum_{j} p_{ij}^{k} \mathbf{x}_j^{k} - \alpha_i^k \nabla f_i (\mathbf{x}_i^k), i \in \mathbb{N}_n, \label{eq:57}
\end{equation}
which defines the classic DGD algorithm~\cite{nedicDistributedSubgradientMethods2009}. 
For more details regarding (\ref{eq:57}), please refer to Appendix~\ref{subsec:Appendix D}.
Howevevr, the update rule (\ref{eq:57}) inherently exhibits an implicit dependence on centralized coordination due to the weighted aggregation term $\sum_{j} p_{ij} \mathbf{x}_j$. This dependence inevitably introduces communication bottlenecks and deviates from the principles of a fully decentralized paradigm, potentially constraining scalability and compromising robustness in large-scale distributed systems.

If implemented with partial-cycle coordination, where each client can start training as long as it receives updates from at least one other client, we refer to the method as Distributed Asynchronous Gradient Descent (DAGD) method~\cite{tsitsiklisDistributedAsynchronousDeterministic1986}. The key feature of this \textit{asynchronous} mechanism is non-blocking updates, where clients can proceed with training as soon as they receive partial information from other clients, without waiting for global synchronization.

Based on the above analysis, it can be observed that when the data distribution across the network is IID, according to (\ref{eq:A1-2}) and (\ref{eq:A1-3}), if FedProx (Distributed PA) or MBGD is used for training, the system will converge to a solution where $\nabla f_i (\mathbf{x}_{i}^{\prime}) = 0$ and $\mu_i^* = 0, \ i \in \mathbb{N}_n$. This indicates that, under ideal conditions, even if each client is trained individually, the global optimal parameters can still be obtained. However, due to the influence of statistical heterogeneity, the data distributions across different clients often exhibit significant differences. 
Therefore, in practical scenarios, data from a single client is generally insufficient to train the global optimal parameters, meaning that the optimal Lagrange multipliers $\mu_i^*$ are typically not all zero, $i \in \mathbb{N}_n$.

According to the Fed-DALD-CC framework, when FedProx terminates, as shown in Fig.~\ref{fig:4}(b), for any client $i$, the parameters will converge to the optimal point  I ($\mathbf{x}_i^{1,*} = \mathbf{x}_i^{\prime\prime}$) of the current outer loop. However, at this stage, only the dual residual $\mathcal{D}^{1,v} = 0$ is guaranteed. To achieve the true global optimum H ($\mathbf{x}_i^{*} = \mathbf{x}_i^{\prime\prime\prime}$), satisfying $\mathcal{C}_i^* = \hat{\mathbf{x}}^* - \mathbf{x}_i^{\prime\prime\prime} = 0, \ i \in \mathbb{N}_n$, FedProx must further update the Lagrange multipliers via (\ref{eq:15}) and continue iterating until condition (\ref{eq:A1-1}) is satisfied. This iterative process leads to FedProx transitioning into the Fed-DALD-CC framework. When the termination condition for the inner layer is set to (B4) and $v_{\text{max}} = 1$, the framework reduces to the consensus-based ADMM framework~\cite{shiLinearConvergenceADMM2014,lingDLMDecentralizedLinearized2015,changMultiagentDistributedOptimization2015,makhdoumiConvergenceRateDistributed2017,caoDifferentiallyPrivateADMM2021,bastianelloAsynchronousDistributedOptimization2021,zhouFederatedLearningInexact2023,zhouFedGiAEfficientHybrid2023,kantFederatedLearningUsing2023}. This progression underscores the significant advantages of the Fed-DALD framework in addressing the challenges posed by statistical heterogeneity.

\subsubsection{Inexact Version}
\label{subsubec:5.2.3Inexact Version}
The DALD framework operates through a three-tiered architecture comprising  an outer layer, an inner layer, and a solver layer. As an accelerated variant of the standard algorithm presented in Section~\ref{sec:3.Fed-DALD}, DALD implements inexact minimization within its inner layer.
Theorem 2 establishes that convergence remains guaranteed even when problem (\ref{eq:5}) is solved inexactly. This raises a critical question: Does DALD preserve convergence under inexact solutions of subproblems in the solver layer? 
Inspired by Theorem 2, it is not difficult to deduce that, under the conditions (B2) or (\ref{eq:29}), DALD remains convergent, even when the subproblems (\ref{eq:13}), (\ref{eq:14}), and (\ref{eq:21}) are solved inexactly within the solver layer.

To facilitate comprehension, we employ FedProx as an illustrative example. As shown in Fig.~\ref{fig:4}(b), when the inexact solution $ \mathbf{x}_i^{\prime\prime\prime\prime} $ (denoted as F) lies to the left of $ \mathbf{x}_i^k $ (denoted as G), i.e., $ \hat{\mathbf{x}}^{k-1} < \mathbf{x}_i^{\prime\prime\prime\prime} < \mathbf{x}_i^k $, solving the surrogate function~(\ref{eq:53}) is equivalent to optimizing its upper bound, with the proximal parameter satisfying $ 0 < \bar{\alpha}_i < \alpha_i $. Conversely, when $ \mathbf{x}_i^{\prime\prime\prime\prime} $ lies to the right of $ \mathbf{x}_i^k $, i.e., $ \mathbf{x}_i^k < \mathbf{x}_i^{\prime\prime\prime\prime}  $, it corresponds to optimizing the lower bound of the surrogate function~(\ref{eq:53}), and the proximal parameter must satisfy $ \underaccent{\bar}{\alpha}_i > \alpha_i $. In FedProx, regardless of the scenario, the solution converges to the optimal point $ \mathbf{x}_i^{\prime\prime} $ through iterative updates, inherently satisfying (\ref{eq:29}).

Next, we consider a more general form of unconstrained optimization problem:
\begin{equation}
\min_{\mathbf{x}} \; f(\mathbf{x}) = \psi(\mathbf{x}) + \pi(\mathbf{x}), \label{eq:58}
\end{equation}
where $ f(\mathbf{x}) $ satisfies Assumption (A1): $ \psi(\mathbf{x}) $ is a continuously differentiable function that explicitly depends on each $ \mathbf{x}_i $ for all $  i \in \mathbb{N}_n $, and both $ f(\mathbf{x}) $ and $ \pi(\mathbf{x}) $ are convex functions, with $ \pi(\mathbf{x}) $ potentially being non-smooth.

When $ \pi(\mathbf{x}) = 0 $, according to Theorem 1, distributed alternating optimization can be directly applied to each component $ \mathbf{x}_i $, reducing the algorithm to the classical BCD method:
\begin{equation}
    \mathbf{x}_i^{k+1} = \arg\min_{\mathbf{x}_i}  \; f_i(\mathbf{x}_i, \mathbf{x}^{k}_{j}, \mathbf{x}^{k+1}_{e}),
\label{eq:59}
\end{equation}
where $ f_i(\cdot) $ denotes the component of $ f(\mathbf{x}) $ that depends on $ \mathbf{x}_i$, and $j, e\in \mathcal{R}_i, j>i,e<i, i \in \mathbb{N}_n $.

If the problem (\ref{eq:58}) is treated as a monolithic optimization problem and solved via GD in the solver layer, the update rule for each element becomes:
\begin{equation}
    \mathbf{x}_i^{k+1} = \mathbf{x}_i^{k} - \alpha \nabla_i f(\mathbf{x}^{k}) = \mathbf{x}_i^{k} - \alpha \nabla_i f_i(\mathbf{x}_i^k, \mathbf{x}_{-i}^k).   \label{eq:60}
\end{equation}

This update rule is equivalent to solving the subproblem:
\begin{equation}
\begin{aligned}
        \mathbf{x}_i^{k+1} = \arg\min_{\mathbf{x}_i}  \nabla_i f_i(\mathbf{x}_i^k, \mathbf{x}_{-i}^k)^{\top}(\mathbf{x}_i - \mathbf{x}_i^k)  +   \frac{1}{2\alpha} \|\mathbf{x}_i - \mathbf{x}_i^k\|^2.
\end{aligned}
\label{eq:61}
\end{equation}

The analytical comparison above reveals that the classical GD method aligns with the Jacobi-type iterative framework in the solver layer. When a partial-cyclic coordination strategy—specifically, the Gauss-Seidel update strategy—is applied to~(\ref{eq:61}),  the classical GD method reduces to the Block Coordinate Gradient Descent (BCGD) method~\cite{beckConvergenceBlockCoordinate2013,wrightCoordinateDescentAlgorithms2015,hongIterationComplexityAnalysis2017,tsengCoordinateGradientDescent2009}. Thus, BCGD can be interpreted as a distributed alternating optimization paradigm of classical GD. 

Given that $ \pi(\mathbf{x}) = 0 $, the exact solution can be obtained by iterating the GD or NM on subproblem~(\ref{eq:59}) in the solver layer until convergence. Notably, any intermediate step of GD iteration inherently satisfies condition~(\ref{eq:29}). Specifically, when the number of local iterations $W$ is set to 1,
the traditional BCD naturally degenerates into the BCGD form.
Furthermore, if $ f(\mathbf{x}) $ is fully separable, i.e., $ f_i(\mathbf{x}_i, \mathbf{x}_{-i}) = f_i(\mathbf{x}_i) $, an interesting observation emerges: Algorithm~(\ref{eq:61}) exhibits structural similarity to~(\ref{eq:47}). The key distinction lies in problem~(\ref{eq:1}), which imposes a consistency constraint on the local parameters of each agent:
$f(\mathbf{x}) = \sum_{i=1}^n f_i(\mathbf{x}),$
whereas problem~(\ref{eq:58}) is formulated as:
$f(\mathbf{x}) = \sum_{i=1}^n f_i(\mathbf{x}_i),$
thereby eliminating the need for a global consensus constraint $ \hat{\mathbf{x}} = \mathbf{x}_i,  i \in \mathbb{N}_n $.

\begin{table*}[t!bp]
  \centering
  \caption{Descriptions of Five Real Datasets}
  \label{tab:1}
  \renewcommand{\arraystretch}{1.25} 
  \setlength{\tabcolsep}{5pt} 
  \footnotesize 
  \begin{tabular}{cccccccc}
    \toprule
    \multirow{2.5}{*}{Datasets} & \multirow{2.5}{*}{Source} & \multirow{2.5}{*}{Instances} & \multirow{2.5}{*}{Features} & \multicolumn{2}{c}{MSE} & \multicolumn{2}{c}{$R^2$ Score} \\
    \cmidrule(lr){5-8}
    & & & & AIO & DALD & AIO & DALD \\
    \midrule
    Diabetes                   & scikit-learn & 442       & 10       & 2859.6963 & 2859.6964 & 0.5177   & 0.5177 \\
    California Housing         & scikit-learn & 20640     & 8        & 0.5243    & 0.5310    & 0.6062   & 0.6012 \\
    Wine Quality               & UCI          & 4898      & 11       & 0.5398    & 0.5407    & 0.2921   & 0.2909 \\
    Abalone                    & UCI          & 4177      & 8        & 4.8027    & 4.8033    & 0.5379   & 0.5378 \\
    Combined Cycle Power Plant & UCI          & 9568      & 4        & 20.7674   & 20.7823   & 0.9287   & 0.9286 \\
    \bottomrule
  \end{tabular}
\end{table*}

When a non-smooth component $ \pi(\mathbf{x}) $ is present, we generalize the approach by constructing a surrogate function $\mathcal{F}$ for $ f $, where $ \mathcal{F} $ serves as an upper bound of $ \psi(\mathbf{x}) $ (or $ f(\mathbf{x}) $). 
This leads to the Block Successive Upper-bound Minimization (BSUM) method~\cite{razaviyaynUnifiedConvergenceAnalysis2013,hongIterationComplexityAnalysis2017}. 
Consequently, the BSUM can be viewed as a variant of BCD that permits inexact solutions to subproblems. 
Specifically, when $ n = 1 $ and a second-order approximation is used to construct the surrogate function for the smooth term $ \psi(\mathbf{x}) $, the BSUM framework reduces to the classical Proximal Gradient (PG) method~\cite[Ch. 6.3]{ bertsekasConvexOptimizationAlgorithms2015},~\cite{beckFastIterativeShrinkagethresholding2009}:
\begin{equation}
\begin{aligned}
    \mathbf{x}^{k+1} =  \arg\min_{\mathbf{x}}
     \; \pi(\mathbf{x}) +  \psi(\mathbf{x}^k)& +   \nabla \psi(\mathbf{x}^k)^\top  (\mathbf{x} - \mathbf{x}^k ) \\ & +   \frac{1}{2\alpha} \|\mathbf{x} - \mathbf{x}^k\|^2.
\end{aligned}
\label{eq:62}
\end{equation}
Simplifying the above expression, we obtain:
\begin{equation}
    \mathbf{x}^{k+1} =\arg\min_{\mathbf{x}} \; \pi(\mathbf{x}) + \frac{1}{2\alpha} \|\mathbf{x} - (\mathbf{x}^k - \alpha \nabla \psi(\mathbf{x}^k))\|^2,\label{eq:63}
\end{equation}
which corresponds to the proximal operator:
\begin{equation}
\mathbf{x}^{k+1} = \operatorname{prox}_{\alpha \pi} \left( \mathbf{x}^k - \alpha \nabla \psi(\mathbf{x}^k) \right). \label{eq:64}
\end{equation}

When extending to $ n \geq 2 $ and applying alternating optimization across blocks, the PG  generalizes to the Block Coordinate Proximal Gradient (BCPG) algorithm~\cite{hongIterationComplexityAnalysis2017}:
\begin{equation}
\mathbf{x}_i^{k+1} = \operatorname{prox}_{\alpha \pi_i} \left( \mathbf{x}_i^k - \alpha \nabla_i \psi_i(\mathbf{x}_i^k, \mathbf{x}_j^k, \mathbf{x}_e^{k+1}) \right), \label{eq:65}
\end{equation}
where $ \psi_i(\cdot) $ and $ \pi_i(\cdot) $ are the components of $ \psi(\mathbf{x}) $ and $ \pi(\mathbf{x}) $ that depend on $ \mathbf{x}_i $, respectively.  Both $ \psi_i(\cdot) $ and $ \pi_i(\cdot) $ are functions of $ \mathbf{x}_i^k, \mathbf{x}_j^k, \mathbf{x}_e^{k+1} $, with $ j, e \in \mathcal{R}_i $, $ j > i $, $ e < i $, $ i \in \mathbb{N}_n $.

\section{Numerical Experiments}
\label{sec:6.Numerical Experiments}
This section presents the numerical results of applying the DALD method to the optimization problem, aiming to further illustrate the proposed approach. For convenience, we set the initial parameters as $\mu^1 = \mathbf{0}$, $\rho = \mathbf{1}$, and the initial solution as $\mathbf{0}$.

\subsection{IID Case: Regression Training}

We begin by applying DALD to linear regression training on five real-world datasets, as presented in Table~\ref{tab:1}. 
For simplification, we assume that all data samples \( D \) are used for training and are evenly distributed across three data nodes. We introduce consensus constraints \( \mathbf{x}_1 = \mathbf{x}_2 \) and \( \mathbf{x}_2 = \mathbf{x}_3 \) to represent linear information routing among the three nodes. The optimization problem can be formulated as:
\begin{equation*}
\min_\mathbf{x} ~\sum_{i=1}^3 \sum_{j=1}^{|D_i|} (a_{ij}^{\top} \mathbf{x}_i - b_{ij})^2
\end{equation*}
\begin{equation*}
\text{s.t.~~} \mathbf{x}_1 = \mathbf{x}_2,  \mathbf{x}_2 = \mathbf{x}_3.
\end{equation*}
Here, \( D_i \) represents the dataset at node \( i \), \( a_{ij} \) denotes the feature vector of the \( j \)-th sample in dataset \( D_i \), and \( b_{ij} \) is the corresponding target value. The variable \( \mathbf{x}_i \) is the local parameters optimized at node \( i \), with the objective of minimizing the loss function while ensuring consistency across the parameters of the nodes.

During distributed training, a subproblem solving sequence $\mathbb{S} = \begin{bmatrix}
1 & 1 & 0 \\
0 & 1 & 1 \\
0 & 0 & 0
\end{bmatrix}$
was employed based on the problem characteristics, and the BFGS method from the \texttt{scipy.optimize} module (version 1.14.1) was used in the solver layer. We applied the \hyperref[condition:B4]{(B4)} stopping criterion, setting \( v_{\max} = 1 \) and terminating the training when the total number of inner loop iterations reached 1000. We then recorded the Mean Squared Error (MSE) and the \( R^2 \) Score.

As shown in Table~\ref{tab:1}, the experimental results illustrate the performance of integrated training (AIO) and distributed training using DALD across multiple datasets, in terms of MSE and \( R^2 \) Score. The results indicate that both methods exhibit similar performance across various test cases. For instance, in the second dataset, the MSE for DALD is 0.5310, slightly higher than AIO’s 0.5243, while the \( R^2 \) Score for DALD is 0.6012, slightly lower than AIO’s 0.6062. Overall, the performances of the two methods are comparable, demonstrating that DALD maintains robustness and consistency in distributed training, achieving stable performance across different nodes and datasets.

\begin{table*}[t!bp]
  \centering
  \caption{Accuracy Mean (\%) and Standard Deviation (\textpertenthousand) Across Different Methods}
  \label{tab:2-accuracy_comparison}
  \renewcommand{\arraystretch}{1.1}
  \setlength{\tabcolsep}{5pt}
  \footnotesize
  \begin{tabular}{cccccccccccccc}
    \toprule
    \multirow{4}{*}{$n$} & \multirow{4}{*}{$\lambda$} & \multicolumn{4}{c}{FedProx} & \multicolumn{4}{c}{Fed-DALD-CC} & \multicolumn{4}{c}{Fed-DALD-DC} \\
    \cmidrule(lr){3-6} \cmidrule(lr){7-10} \cmidrule(lr){11-14}
    & & \multicolumn{2}{c}{$\text{Iters} = 1000$} & \multicolumn{2}{c}{$\text{Iters} = 3000$} & \multicolumn{2}{c}{$\text{Iters} = 1000$} & \multicolumn{2}{c}{$\text{Iters} = 3000$} & \multicolumn{2}{c}{$\text{Iters} = 1000$} & \multicolumn{2}{c}{$\text{Iters} = 3000$} \\
    \cmidrule(lr){3-4} \cmidrule(lr){5-6} \cmidrule(lr){7-8} \cmidrule(lr){9-10} \cmidrule(lr){11-12} \cmidrule(lr){13-14}
    & & Mean & Std Dev & Mean & Std Dev & Mean & Std Dev & Mean & Std Dev & Mean & Std Dev & Mean & Std Dev \\
    \midrule
    \multirow{4}{*}{10} & 0       & 97.43 & 27.11 & 97.42 & 31.35 & 97.44 & 15.66 & 97.95 & 5.87  & 98.80 & 2.14  & 99.04 & 1.50  \\ 
                        & $10^{-4}$ & 97.42 & 27.21 & 97.41 & 32.28 & 97.43 & 15.71 & 97.95 & 5.96  & 98.80 & 2.14  & 99.03 & 1.82  \\
                        & $10^{-3}$ & 97.41 & 28.23 & 97.40 & 32.12 & 97.42 & 15.61 & 97.94 & 6.55  & 98.78 & 2.23  & 99.01 & 1.65  \\
                        & $10^{-2}$ & 97.23 & 26.88 & 97.16 & 28.65 & 97.31 & 17.01 & 97.79 & 10.13 & 98.57 & 2.77  & 98.64 & 1.69  \\
    \midrule
    \multirow{4}{*}{50} & 0       & 95.81 & 93.98 & 95.87 & 93.95 & 98.08 & 4.69  & 98.15 & 1.81  & 98.40 & 6.26  & 98.73 & 2.51  \\
                        & $10^{-4}$ & 95.81 & 93.62 & 95.86 & 94.01 & 98.07 & 4.83  & 98.15 & 1.96  & 98.40 & 6.18  & 98.73 & 2.48  \\
                        & $10^{-3}$ & 95.75 & 94.89 & 95.74 & 96.03 & 98.02 & 4.79  & 98.10 & 2.16  & 98.37 & 5.67  & 98.67 & 1.91  \\
                        & $10^{-2}$ & 95.12 & 101.87 & 94.97 & 107.57 & 97.51 & 13.66 & 97.69 & 2.34  & 97.80 & 9.71  & 97.90 & 3.16  \\
    \midrule
    \multirow{4}{*}{100} & 0      & 94.60 & 141.70 & 94.72 & 139.52 & 98.04 & 3.29  & 98.13 & 2.22  & 98.05 & 6.52  & 98.49 & 6.35  \\
                        & $10^{-4}$ & 94.59 & 141.79 & 94.69 & 139.88 & 98.03 & 3.29  & 98.12 & 2.17  & 98.04 & 6.54  & 98.48 & 6.18  \\
                        & $10^{-3}$ & 94.45 & 143.76 & 94.44 & 143.74 & 97.96 & 3.24  & 98.04 & 1.78  & 97.99 & 6.57  & 98.38 & 5.37  \\
                        & $10^{-2}$ & 93.34 & 149.71 & 93.15 & 153.74 & 97.21 & 5.89  & 97.26 & 2.46  & 97.23 & 8.34  & 97.29 & 4.16  \\
    \midrule
    \multirow{4}{*}{200} & 0      & 92.92 & 247.31 & 93.32 & 234.14 & 97.90 & 3.89  & 98.15 & 2.42  & 97.75 & 9.35  & 98.24 & 5.87  \\
                        & $10^{-4}$ & 92.89 & 248.07 & 93.25 & 235.36 & 97.89 & 3.94  & 98.13 & 2.65  & 97.75 & 9.48  & 98.23 & 5.84  \\
                        & $10^{-3}$ & 92.65 & 251.73 & 92.79 & 242.19 & 97.79 & 4.49  & 98.00 & 2.55  & 97.66 & 10.07 & 98.07 & 7.96  \\
                        & $10^{-2}$ & 90.42 & 289.08 & 90.20 & 295.19 & 96.78 & 5.74  & 96.72 & 2.62  & 96.71 & 9.66  & 96.74 & 4.67  \\
    \midrule
    \multirow{4}{*}{500} & 0      & 88.58 & 490.55 & 90.71 & 389.28 & 97.54 & 7.01  & 97.99 & 2.58  & 97.32 & 15.59 & 97.86 & 9.85  \\
                        & $10^{-4}$ & 88.49 & 494.32 & 90.45 & 401.05 & 97.52 & 7.16  & 97.96 & 2.36  & 97.30 & 15.68 & 97.84 & 9.78  \\
                        & $10^{-3}$ & 87.82 & 519.60 & 88.92 & 467.79 & 97.32 & 7.43  & 97.67 & 3.33  & 97.14 & 15.83 & 97.55 & 11.22 \\
                        & $10^{-2}$ & 81.56 & 700.42 & 82.40 & 680.71 & 96.10 & 6.72  & 96.02 & 2.16  & 95.96 & 22.54 & 95.96 & 13.49 \\
    \midrule
    \multirow{4}{*}{1000} & 0     & 86.68 & 616.80 & 91.96 & 367.89 & 97.29 & 7.16  & 97.81 & 3.07  & 97.00 & 18.80 & 97.53 & 13.89 \\
                        & $10^{-4}$ & 86.67 & 621.43 & 91.17 & 384.79 & 97.26 & 7.22  & 97.76 & 2.94  & 96.98 & 18.82 & 97.49 & 13.76 \\
                        & $10^{-3}$ & 85.31 & 649.78 & 88.21 & 507.81 & 97.01 & 7.44  & 97.26 & 3.51  & 96.86 & 19.59 & 97.08 & 10.93 \\
                        & $10^{-2}$ & 74.46 & 879.23 & 76.85 & 845.36 & 95.54 & 10.65 & 95.49 & 3.06  & 95.24 & 52.64 & 95.29 & 35.06 \\
    \bottomrule
  \end{tabular}
\end{table*}

\subsection{ Non-IID Case: Classification Training}

This section constructs a binary classification task based on the MNIST dataset, aiming to distinguish handwritten digits 3 and 7 using a logistic regression model to evaluate algorithm performance. Each \( 28 \times 28 \) pixel image is vectorized into a 784-dimensional feature, with the complete dataset containing 12,396 training samples. To simulate the non-IID data scenario in a distributed learning environment, we employ a stratified sampling strategy to partition the data across \( n \) clients, implemented as follows:  
\begin{itemize}
    \item[a)] Even-indexed clients: Allocated a 4:1 class ratio (digit 3:7) by selecting samples with strides \(n\)  and \(4n\).
    \item[b)] Odd-indexed clients: Assigned an inverse 1:4 ratio (digit 3:7) using strides \(4n\)  and \(n\).
\end{itemize}

Consider the following \( \ell_1 \)-regularized optimization problem:
\begin{equation*}
\min_{\mathbf{x}}  \sum_{i=1}^{n} f_i(\mathbf{x}_i) + n\lambda \|\mathbf{x}_i\|_1, \text{~s.t.~} \mathcal{C}= 0,
\end{equation*}
where the regularization parameter \( n\lambda \) controls model sparsity, and the consensus constraint \( \mathcal{C} \) ensures parameter consistency. The local objective function is defined as:
\begin{equation*}
f_i(\mathbf{x}_i) = \frac{1}{N} \sum_{j=1}^{|D_i|} \ln\left[ 1 + \exp\left( -b_{ij} (a_{ij}^{\top} \mathbf{x}_i) \right) \right] + \lambda \|\mathbf{x}_i\|_1,
\end{equation*}
where \( a_{ij} \in \mathbb{R}^{784} \) represents the feature, \( b_{ij} \in \{-1,+1\} \) denotes the label in dataset \( D_i \), and \( \mathbf{x}_i \) is the local parameter to be optimized at client \( i \), $i \in \mathbb{N}_n$. Under the DALD framework, the local augmented Lagrangian is expressed as:
$$\begin{aligned}
   \Lambda_{\rho_i}^i (\mathbf{x}_i, \mathbf{x}_j^{k,v-1},  \mathbf{x}_e^{k,v}, &\;\mu^k ) 
= f_i (\mathbf{x}_i ) + \mathcal{A}_{\rho_i}^i (\mathbf{x}_i, \mathbf{x}_j^{k,v-1}, \mathbf{x}_e^{k,v}, \mu_i^k ) \\
&~= \Psi_i (\mathbf{x}_i, \mathbf{x}_j^{k,v-1}, \mathbf{x}_e^{k,v}, \mu_i^k, \rho_i ) + \Pi_i (\mathbf{x}_i).
\end{aligned}$$
where \( \Pi_i (\mathbf{x}_i ) = \lambda \|\mathbf{x}_i\|_1 \), corresponding to (\ref{eq:12}) and (\ref{eq:20}). In the DALD framework, an inexact solution strategy is adopted for both the inner layer and solver layer. Specifically, in the inner layer, condition (B4) is applied with \( v_{\max} = 1 \), while in the solver layer, the BCPG algorithm (\ref{eq:65}) is employed  with $W=1$ as the local iteration limit, using $\mathbf{x}_i^{k,v,0}=\mathbf{x}_i^{k,v-1}$ as the initialization.
Since \( \Pi_i (\mathbf{x}_i ) = \lambda \|\mathbf{x}_i\|_1 \), the proximal operator (\ref{eq:65}) reduces to the soft-thresholding operator \cite{Tibshirani2019}. This yields the closed-form update:
\begin{equation}
\begin{aligned}
\mathbf{x}_i^{k,v,w} ={}& \mathcal{S}_{\lambda\alpha} \left( \mathbf{x}_i^{k,v,w-1} \right. \\
&\left. \; - \alpha \nabla_i \Psi_i ( \mathbf{x}_i^{k,v,w-1}, \mathbf{x}_j^{k,v-1}, \mathbf{x}_e^{k,v}, \mu_i^k, \rho_i  ) \right),
\end{aligned}
\label{eq:66}
\end{equation}
where \( \mathcal{S}_{\lambda\alpha} (\cdot) \) is defined component-wise as: $[\mathcal{S}_{\lambda\alpha} (\beta)]_r = \operatorname{sign}(\beta_r)  \max (|\beta_r| - \lambda\alpha, 0), r\in\mathbb{N}_{|\beta |}.$

In this case, FedProx is selected as the baseline method to validate the effectiveness of distributed optimization algorithms in non-IID data scenarios by comparing it with the proposed Fed-DALD-CC and Fed-DALD-DC algorithms. Specifically, DALD-DC serializes subproblem solving by imposing linear routing constraints between adjacent nodes: $\mathbf{x}_i - \mathbf{x}_{i+1} = 0,  i \in \mathbb{N}_{n-1}.$
The experiments adopt a unified convergence threshold of $\epsilon_{\text{dual}} = \epsilon_{\text{pri}} = 1 \times 10^{-5}$, with  a constant step size $\alpha=10^{-4}$.  The maximum number of inner loops set to 1000 and 3000, respectively. All methods are evaluated under varying numbers of clients $n \in \{10, 50, 100, 200, 500, 1000\}$ and regularization parameters $\lambda \in \{0, 10^{-4}, 10^{-3}, 10^{-2}\}$, where $\lambda=0$ denotes the non-regularized baseline. The results are summarized in Table~\ref{tab:2-accuracy_comparison}.

\begin{figure*}[t]
    \centering
    \includegraphics[width=0.97\linewidth, trim=0cm 0cm 0.2cm 0cm]{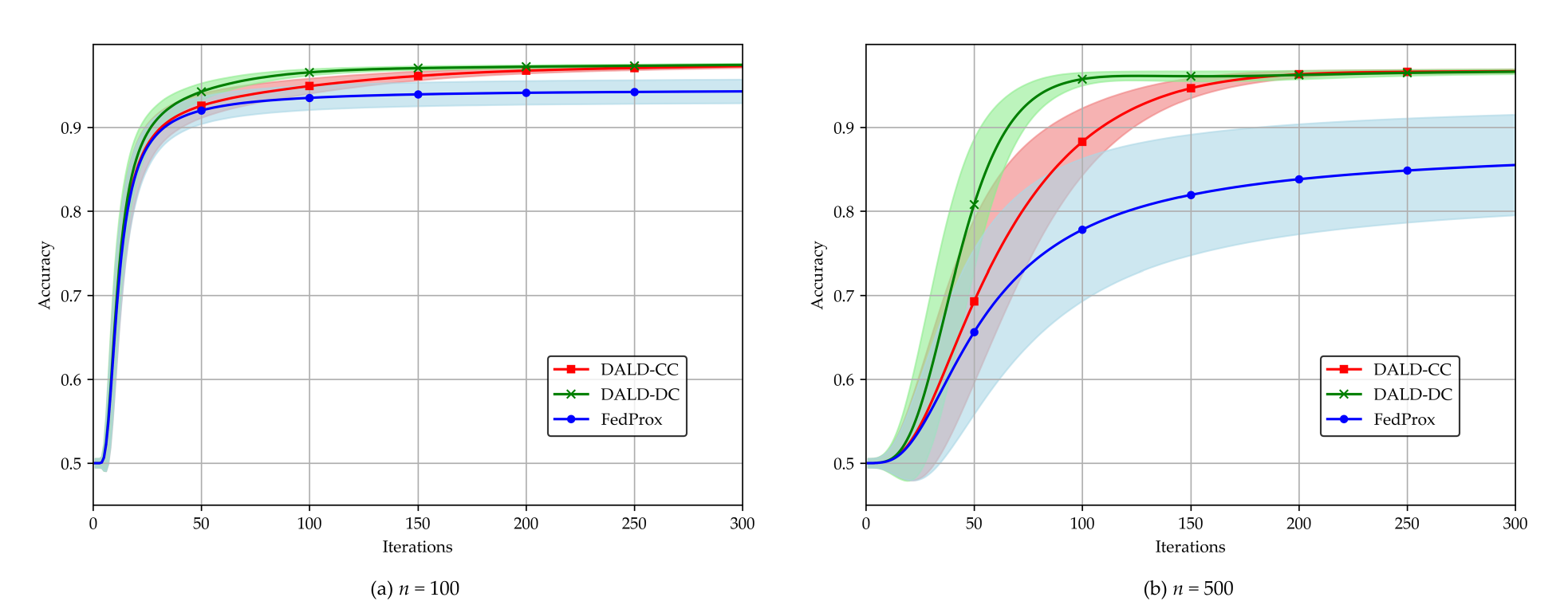}  
    \caption{Performance Comparison of Algorithms at Varying Client Scales with $\lambda=10^{-3}$}
    \label{fig:5}
\end{figure*}

Taking $\lambda = 10^{-3}$ as an example, Table~\ref{tab:2-accuracy_comparison} demonstrates that, under the same number of iterations, the DALD algorithms significantly outperform FedProx. When the number of clients is $n = 1000$ and the number of iterations reaches 3000, DALD-DC achieves an accuracy of 97.08\%, representing an 9.87 percentage point improvement over FedProx’s 87.21\%. As the number of clients increases from $n = 10$ to $n = 1000$, the accuracy of FedProx drops by 10.19\% (from 97.40\% to 87.21\%), whereas DALD-DC exhibits only a 1.93\% decrease (from 99.01\% to 97.08\%), indicating the superior adaptability of the proposed method in large-scale distributed scenarios.

Regarding stability, the standard deviation of FedProx increases from 32.12 at $n=10$ to 507.81 at $n=1000$, highlighting its limitations in handling statistical heterogeneity. In contrast, DALD-DC maintains a standard deviation of only 10.93 under the same conditions, demonstrating its effectiveness in suppressing performance fluctuations caused by statistical heterogeneity. Notably, as the number of clients grows, statistical heterogeneity becomes more pronounced. Table~\ref{tab:2-accuracy_comparison} reveals that for $n=500$ and $n=1000$, FedProx exhibits an order-of-magnitude increase in standard deviation, leading to significant performance degradation, whereas DALD-CC and DALD-DC maintain consistently low standard deviations with accuracy exceeding 95\%, showcasing superior stability and generalization capability.

Theoretical analysis in section~\ref{subec:5.2Framework Unification} further reveals that the DALD algorithms mitigate the adverse effects of statistical heterogeneity in FL by optimizing the Lagrange multipliers, ensuring stable convergence in large-scale heterogeneous data scenarios. To intuitively illustrate this conclusion, Fig.~\ref{fig:5} compares the performance of the three algorithms over the first 300 iterations for $n=100$ and $n=500$. The results indicate that, under the same maximum iteration constraint, the accuracy and stability of FedProx consistently lag behind those of the two DALD algorithms.

The experimental results empirically validate the superiority of the proposed method in handling non-IID data and overcoming statistical heterogeneity. Particularly in large-scale distributed environments, their performance advantages become more pronounced, offering an effective solution for FL applications in complex real-world scenarios.

\section{Conclusion}\label{sec:7.Conclusion}
This paper proposes a distributed optimization framework, Fed-DALD, designed to address large-scale FL tasks characterized by statistical heterogeneity and privacy constraints. To mitigate prohibitive computational in initial phases, accelerated algorithm variants are developed with formally guaranteed convergence properties. During the algorithmic design, we introduce a strategy that incorporates proximal relaxation and second-order approximation to construct surrogate functions for the original objective, which are then optimized in an alternating manner. Within this framework, we systematically derive multiple classes of classical unconstrained optimization algorithms, bridging theoretical gaps among existing methods. This unification is expected to provide a novel theoretical perspective and a cohesive analytical foundation to the optimization community. Future research will focus on exploring its adaptability to various communication topologies and its extension to asynchronous computing paradigms.

\section{Appendices}
\label{sec:8.Appendices}
\subsection{Computational Procedure for   \( \mathcal{D}^{\textit{k,v}} \) in  Fed-DALD-CC}
\label{subsec:appendix A}
The necessary and sufficient optimality conditions for problem (\ref{eq:8}) consist of primal feasibility,
\begin{equation}
\hat{\mathbf{x}}^* - \mathbf{x}_i^* = 0, i \in \mathbb{N}_n
\label{eq:A1-1}
\end{equation}
and dual feasibility derived from the Lagrangian (\ref{eq:9}),
\begin{equation}
0 \in \partial_i f_i(\mathbf{x}_i^*) - \mu_i^*, i \in \mathbb{N}_n, 
\label{eq:A1-2}
\end{equation}
\begin{equation}
\hat{\mu}^*=\sum_{i=1}^{n} \mu_i^* = 0.
\label{eq:A1-3}
\end{equation}
As \( v \to \infty \),  \( \hat{\mathbf{x}}^{k,v} \) minimizes the problem (14). We have that
\begin{equation}
\sum_{i=1}^{n} \mu_i^k + 2\sum_{i=1}^{n} \rho_i \circ \rho_i \circ \mathcal{C}_i^{k,v} = \sum_{i=1}^{n} \mu_i^{k+1} = 0, 
\label{eq:A1-4}
\end{equation}
which means that \( \mu_i^{k+1} \) always satisfies (\ref{eq:A1-3}), \( i \in \mathbb{N}_n \).
When \( v \to \infty \), let \( \mathbf{x}_i^{k,v} \) minimize the problem (\ref{eq:13}), \( i \in \mathbb{N}_n \), we have that
\begin{equation}
0 \in \partial_i f_i(\mathbf{x}_i^{k,v}) - \mu_i^{k+1} + 2\rho_i \circ \rho_i \circ (\hat{\mathbf{x}}^{k,v} - \hat{\mathbf{x}}^{k,v-1}). 
\label{eq:A1-5}
\end{equation}
As \( v \to \infty \), we obtain the limit point \( (\bar{\mathbf{x}}^{k,v}, \bar{\hat{\mathbf{x}}}^{k,v}) \) at the \( k \)-th outer loop. According to Theorem 1, this point is also a stationary point, resulting
\begin{equation}
\mathcal{D}^{k,v} = \hat{\mathbf{x}}^{k,v} - \hat{\mathbf{x}}^{k,v-1} = 0. 
\label{eq:A1-6}
\end{equation}
When \( k \to \infty \), we have \( (\bar{\mathbf{x}}^{k,v}, \bar{\hat{\mathbf{x}}}^{k,v}) \to (\mathbf{x}^*, \hat{\mathbf{x}}^*) \) and \( \bar{\mu}^k \to \mu^* \). Combining (\ref{eq:A1-5}) and (\ref{eq:A1-6}), this ultimately satisfies  (\ref{eq:A1-2}).
We will refer to
\begin{equation}
\mathcal{D}^k =  \bar{\hat{\mathbf{x}}}^{k,v} - \bar{\hat{\mathbf{x}}}^{k,v-1}
\label{eq:A1-7}
\end{equation}
as the \textit{dual residual} and to
\begin{equation}
\mathcal{C}_i^k  = \bar{\hat{\mathbf{x}}}^{k,v} - \bar{\mathbf{x}}_i^{k,v}, \label{eq:A1-8}
\end{equation}
as the \textit{primal residual} at outer loop \( k \), \( i \in \mathbb{N}_n \).

\subsection{Computational Procedure for  \( \mathcal{D}^{\textit{k,v}} \) in  Fed-DALD-DC}

\label{subsec:Appendix B}
The necessary and sufficient optimality conditions for problem (\ref{eq:16}) consist of primal feasibility,
\begin{equation}
\mathbf{x}_i^* - \mathbf{x}_j^* = 0,  i \in \mathbb{N}_n, \, j \in \mathcal{R}_i, \, j > i,
\label{eq:A2-1}
\end{equation}
and dual feasibility derived from the Lagrangian (\ref{eq:17}),
\begin{equation}
0 \in \partial_i f_i(\mathbf{x}_i^*) + \sum_{j \in \mathcal{R}_i, j > i} \mu_{ij}^* - \sum_{e \in \mathcal{R}_i, e < i} \mu_{ei}^*, i \in \mathbb{N}_n.
\label{eq:A2-2}
\end{equation}
Let \( \mathbf{x}_i^{k,v} \) minimize the problem (\ref{eq:21}), where \( i \in \mathbb{N}_n \), \( j, e \in \mathcal{R}_i \), \( j > i \),  \( e < i \). We obtain
\begin{equation}
0 \in \partial_i \Lambda^i_{\rho_i} \left( \mathbf{x}_i^{k,v}, \mathbf{x}_j^{k,v-1}, \mathbf{x}_e^{k,v}, \mu^k_i \right) = \partial_i f_i(\mathbf{x}_i) + \mathcal{V}_i, 
\label{eq:A2-3}
\end{equation}
where $\mathcal{V}_i$
\begin{equation*}
    \begin{aligned}
         = ~ & \sum_{j \in \mathcal{R}_i, j > i} \left[ \mu_{ij}^k + 2\rho_{ij} \circ \rho_{ij} \circ \left( \mathbf{x}_{ij}^{k,v} - \mathbf{x}_{ji}^{k,v-1} \right) \right]   \\
        & ~ - \sum_{e \in \mathcal{R}_i, e < i} \left[ \mu_{ei}^k + 2\rho_{ei} \circ \rho_{ei} \circ \left( \mathbf{x}_{ei}^{k,v} - \mathbf{x}_{ie}^{k,v} \right) \right]. \\
    \end{aligned}
\end{equation*}
As $v \to \infty$, $\mathcal{V}_i$
\begin{equation*}
    \begin{aligned}
        = ~ & \sum_{j \in \mathcal{R}_i, j > i} \left[ \mu_{ij}^k + 2\rho_{ij} \circ \rho_{ij} \circ \left( \mathbf{x}_{ij}^{k,v} - \mathbf{x}_{ji}^{k,v} + \mathbf{x}_{ji}^{k,v} \right.\right. \\
        & ~ \left. \left. -~ \mathbf{x}_{ji}^{k,v-1}\right)\right]  - \sum_{e \in \mathcal{R}_i, e < i} \mu_{ei}^{k+1} \\
        = ~ & \sum_{j \in \mathcal{R}_i, j > i} \left[\mu_{ij}^{k+1} + 2\rho_{ij} \circ \rho_{ij} \circ \left( \mathbf{x}_{ji}^{k,v} - \mathbf{x}_{ji}^{k,v-1}\right)\right]\\
        & ~- \sum_{e \in \mathcal{R}_i, e < i} \mu_{ei}^{k+1}.
    \end{aligned}
\end{equation*}
When \( v \to \infty \), a stable limit point \( \bar{\mathbf{x}}^{k,v} \) is obtained. This implies that for any subproblem \( i \), there exists: $2\rho_{ij} \circ \rho_{ij} \circ \left( \mathbf{x}_{ji}^{k,v} - \mathbf{x}_{ji}^{k,v-1} \right) = 0,  i \in \mathbb{N}_n, \, j \in \mathcal{R}_i, \, j > i$.
This leads to the following condition:
\begin{equation}
\mathcal{D}_{ij}^{k,v} = \mathbf{x}_{ji}^{k,v} - \mathbf{x}_{ji}^{k,v-1},  i \in \mathbb{N}_n, \, j \in \mathcal{R}_i, \, j > i, 
\label{eq:A2-4}
\end{equation}
which can be interpreted as a residual for the dual feasibility condition (\ref{eq:A2-2}) during the \((k,v)\)-th loop iteration. As \( v \to \infty \) at outer loop \( k \), we obtain
\begin{equation}
\mathcal{D}_{ij}^{k,v} = 0,  i \in \mathbb{N}_n, \, j \in \mathcal{R}_i, \, j > i, 
\label{eq:A2-5}
\end{equation}
which ensures that condition (\ref{eq:A2-2}) is satisfied.
We will refer to 
\begin{equation}
\mathcal{D}_{ij}^{k} =  \bar{\mathbf{x}}_{ji}^{k,v} - \bar{\mathbf{x}}_{ji}^{k,v-1}
\label{eq:A2-6}
\end{equation}
as the \textit{dual residual} and to
\begin{equation}
\mathcal{C}_{ij}^k =  \bar{\mathbf{x}}_{ij}^{k,v} - \bar{\mathbf{x}}_{ji}^{k,v},
\label{eq:A2-7}
\end{equation}
as the \textit{primal residual} at outer loop \( k \),  \( i \in \mathbb{N}_n \), \( j \in \mathcal{R}_i \),  \( j > i \).

Moreover, from the primal variable update rule of Fed-DALD-DC,  definition (\ref{eq:A2-4}) can be equivalently written as
\begin{equation}
    \mathcal{D}^{k,v} = \left\{ \mathcal{D}_i^k = \mathbf{x}_i^{k,v} - \mathbf{x}_i^{k,v-1} \mid i \in \mathbb{N}_n \setminus \{1\} \right\},
    \label{eq:A2-8}
\end{equation}
where \(\{1\}\) (with a slight abuse of notation) represents the level identifier 1, corresponding to the bottom level in the hierarchical network. This also aligns with (\ref{eq:A1-6}).

\subsection{Proof of Lemma 1}
\label{subsec:Appendix C}
\textbf{\textit{Proof:}} (Necessity Proof) For any \( \delta \in (0,1) \), the point \( \mathbf{x}_\delta = (1 - \delta) \mathbf{x}^{k,*} + \delta \mathbf{x} \) $\in$ \( \mathbb{R}^{mn} \). Furthermore, from the optimality of \( \mathbf{x}^{k,*} \), for sufficiently small \( \delta \), we have
\[
\Psi (\mathbf{x}_\delta) + \Pi (\mathbf{x}_\delta) \geq \Psi (\mathbf{x}^{k,*}) + \Pi (\mathbf{x}^{k,*}).
\]
This can be rewritten as
\begin{equation*}
\begin{aligned}
  \Psi \left( (1 - \delta) \mathbf{x}^{k,*} + \delta \mathbf{x} \right) + \Pi \left( (1 - \delta) \mathbf{x}^{k,*} + \delta \mathbf{x} \right)  \\ 
   \geq \Psi (\mathbf{x}^{k,*}) + \Pi (\mathbf{x}^{k,*}).
\end{aligned}
\end{equation*}
Using the convexity of \( \Pi (\mathbf{x}) \), we obtain
\begin{equation*}
\begin{aligned}
  \Psi \left( (1 - \delta) \mathbf{x}^{k,*} + \delta \mathbf{x} \right) + (1 - \delta) \Pi (\mathbf{x}^{k,*}) + \delta \Pi (\mathbf{x})  \\ 
   \geq \Psi (\mathbf{x}^{k,*}) + \Pi (\mathbf{x}^{k,*}),
\end{aligned}
\end{equation*}
which can be further rewritten as
\begin{equation*}
\begin{aligned}
  \frac{\Psi \left( \mathbf{x}^{k,*} + \delta (\mathbf{x} - \mathbf{x}^{k,*}) \right) - \Psi (\mathbf{x}^{k,*})}{\delta} 
   \geq \Pi (\mathbf{x}^{k,*}) - \Pi (\mathbf{x}).
\end{aligned}
\end{equation*}
Letting \( \delta \to 0^+ \), and utilizing the differentiability of \( \Psi (\mathbf{x}) \), we get
\[
\Psi^{\prime} (\mathbf{x}^{k,*}; \mathbf{x} - \mathbf{x}^{k,*}) \geq \Pi (\mathbf{x}^{k,*}) - \Pi (\mathbf{x}),
\]
where \( \Psi^{\prime} (\mathbf{x}^{k,*}; \mathbf{x} - \mathbf{x}^{k,*}) \) denotes the directional derivative of \( \Psi (\mathbf{x}) \) in the direction of \( \mathbf{x} - \mathbf{x}^{k,*} \)
:
\begin{equation*}
\fontsize{11}{11}\selectfont
\begin{aligned}
  &=  \lim_{\delta \to 0^+} \frac{\Psi (\mathbf{x}^{k,*} + \delta (\mathbf{x} - \mathbf{x}^{k,*})) - \Psi (\mathbf{x}^{k,*})}{\delta}  \\ 
    &=   \lim_{\delta \to 0^+} \frac{\Psi (\mathbf{x}^{k,*}) + \delta \langle \nabla \Psi (\mathbf{x}^{k,*}), \mathbf{x} - \mathbf{x}^{k,*} \rangle - \Psi (\mathbf{x}^{k,*})}{\delta} \\
    & =  \langle \nabla \Psi (\mathbf{x}^{k,*}), \mathbf{x} - \mathbf{x}^{k,*} \rangle.
\end{aligned}
\end{equation*}
Therefore, for any \( \mathbf{x} \in \mathbb{R}^{mn} \), we have
\[
\Pi (\mathbf{x}) \geq \Pi (\mathbf{x}^{k,*}) + \langle - \nabla \Psi (\mathbf{x}^{k,*}), \mathbf{x} - \mathbf{x}^{k,*} \rangle,
\]
which implies that \( - \nabla \Psi (\mathbf{x}^{k,*}) \in \partial \Pi (\mathbf{x}^{k,*}) \). In other words, we can use the differentiable function $\Psi$ to characterize the subdifferential of $\Pi$ in any direction.

(Sufficiency Proof) If $-\nabla \Psi (\mathbf{x}^{k,*}) \in \partial \Pi (\mathbf{x}^{k,*})$, we have
\[
0 \in \nabla \Psi (\mathbf{x}^{k,*}) + \partial \Pi (\mathbf{x}^{k,*}) = \partial \mathcal{F} (\mathbf{x}^{k,*}).
\]
According to Definition 1, it follows that $\mathbf{x}^{k,*}$ is a stationary point of the problem $\mathcal{F} (\mathbf{x})$. Utilizing the fact that \( \mathcal{F}(\mathbf{x}) \) is convex, it can be concluded that $\mathbf{x}^{k,*}$ is an optimal solution. 

\subsection{Derivation of the Update Rule~(\ref{eq:57})}
\label{subsec:Appendix D}
Given that $ f $ is a continuously differentiable convex function, the gradient of the surrogate function $\mathcal{F}$ defined in (\ref{eq:53}) can be expressed component-wise as:
\begin{equation*}
    \nabla_{\hat{\mathbf{x}}} \mathcal{F} = \hat{\mathbf{x}} - \sum p_i \mathbf{x}_i,
\end{equation*}
\begin{equation*}
    \nabla_i \mathcal{F} = \nabla_i f_i(\mathbf{x}_i) + \frac{1}{\alpha_i} (\mathbf{x}_i - \hat{\mathbf{x}}),  i \in \mathbb{N}_n.
\end{equation*}
Following Theorem 1, we alternately optimize with respect to $\mathbf{x}_i$ and $\hat{\mathbf{x}}$.
Owing to the differentiability of $\mathcal{F}$, GD can be directly employed to solve the optimization problems.
This yields the following update for $\mathbf{x}_i$: 
\begin{equation}
    \mathbf{x}_i^{k, v+1} = \mathbf{x}_i^{k,v} - \alpha_i^k \nabla_i \mathcal{F} = \hat{\mathbf{x}}^k - \alpha_i^k \nabla f_i(\mathbf{x}_i^{k,v}),  i \in \mathbb{N}_n.\label{eq:A4-1}
\end{equation}
The update for $\hat{\mathbf{x}}$ is given by (\ref{eq:55}). For (\ref{eq:A4-1}), if the number of iterations in the solver layer is limited to a single step, 
then substituting (\ref{eq:55}) into (\ref{eq:A4-1}) directly yields (\ref{eq:57}). As discussed in Section~\ref{subsubec:5.2.3Inexact Version}, the convergence of update rule (\ref{eq:57}) can be rigorously established under the stated assumptions. This concludes the derivation.



\bibliography{references.bib} 

\bibliographystyle{IEEEtran}

\begin{IEEEbiography}[{\includegraphics[width=1in,height=1.25in,clip,keepaspectratio]{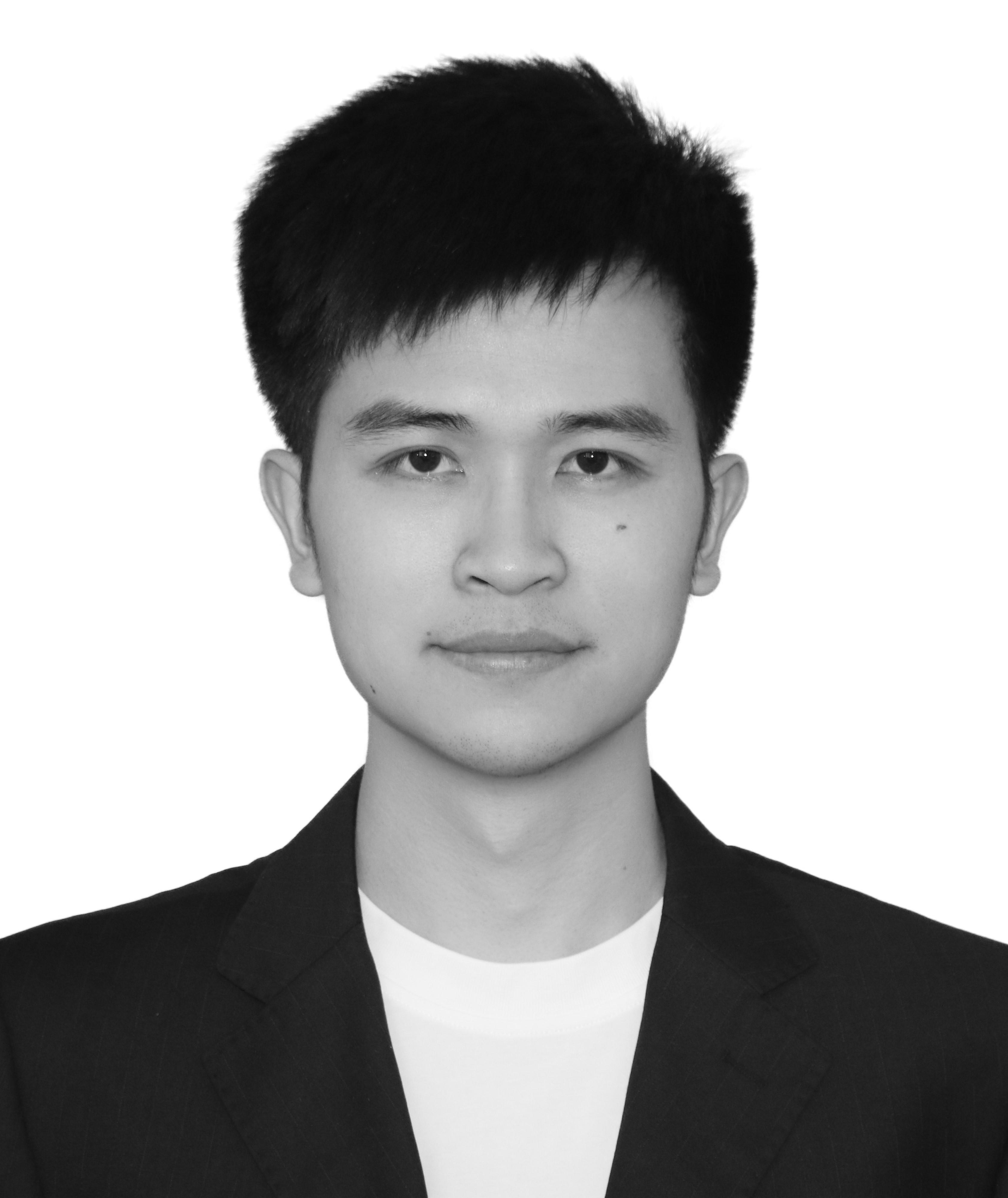}}]{Wenyou Guo}  received his B.Eng. and M.Eng. degrees  in Industrial Engineering from Jiangxi University of Science and Technology, Ganzhou, China, in 2019 and 2022, respectively. He is currently pursuing a Ph.D. degree in Management Science and Engineering at Jinan University, Guangzhou, China. His current research interests include distributed optimization, federated learning, blockchain, and intelligent manufacturing.
\end{IEEEbiography}
\begin{IEEEbiography}
[{\includegraphics[width=1in,height=1.25in,clip,keepaspectratio]{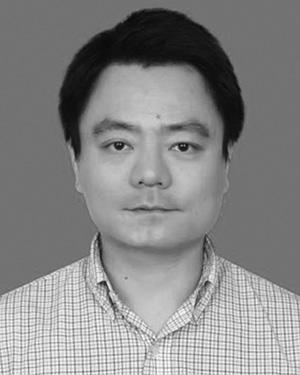}}]{Ting Qu}  received the B.Eng. and M.Phil. degrees in Mechanical Engineering from Xi’an Jiaotong University, Xi’an, China, in 2001 and 2004, respectively, and the Ph.D. degree in Industrial and Manufacturing Systems Engineering from The University of Hong Kong, Hong Kong, in 2008.

He is currently a Full Professor at the School of Intelligent Systems Science and Engineering, Jinan University (Zhuhai Campus), Zhuhai, China. He has undertaken over 20 research projects funded by government and industry, and has published nearly 200 technical papers, approximately half of which have appeared in leading international journals. His research interests include IoT-enabled smart manufacturing systems, logistics and supply chain management, and production service systems.

Dr. Qu serves as a director or board member of several academic associations in the fields of industrial engineering and smart manufacturing.
\end{IEEEbiography}
\begin{IEEEbiography}[{\includegraphics[width=1in,height=1.25in,clip,keepaspectratio]{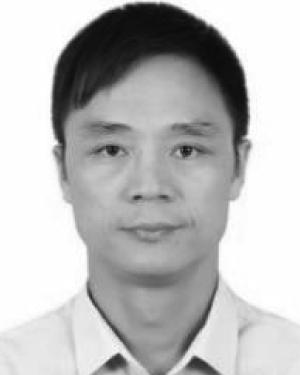}}]{Chunrong Pan}   received the M.S. degree in Mechatronics Engineering from Shantou University, Shantou, China, in 2006, and the Ph.D. degree in Mechanical Engineering from Guangdong University of Technology, Guangzhou, China, in 2010.

From 1997 to 2011, he was affiliated with Shantou University. Since 2011, he has been with Jiangxi University of Science and Technology, Ganzhou, China, where he is currently a Professor in the School of Mechanical and Electrical Engineering. He was a Visiting Scholar at the New Jersey Institute of Technology, Newark, NJ, USA, from 2013 to 2014, and at Bournemouth University, Poole, U.K., from 2018 to 2019. He has over 90 publications, including one book. His research interests include manufacturing system modeling and scheduling, Petri nets, and discrete event systems.
\end{IEEEbiography}
\begin{IEEEbiography}
[{\includegraphics[width=1in,height=1.25in,clip,keepaspectratio]{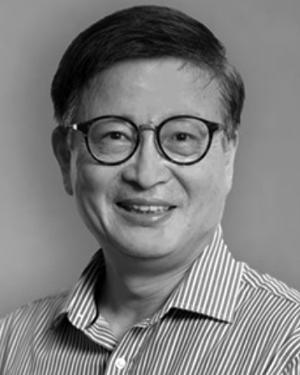}}]{George Q. Huang}  received the B.Eng. degree in Mechanical Engineering from Southeast University, Nanjing, China, in 1983, and the Ph.D. degree in Mechanical Engineering from Cardiff University, Cardiff, U.K., in 1991.

He joined the Department of Industrial and Systems Engineering at The Hong Kong Polytechnic University, Hong Kong, in December 2022 as a Chair Professor of Smart Manufacturing.
Prior to this appointment, he was a Chair Professor of Industrial and Systems Engineering and Head of the Department in the Department of Industrial and Manufacturing Systems Engineering at The University of Hong Kong, Hong Kong.
He has conducted research projects in the areas of smart manufacturing, logistics, and construction, with a focus on IoT-enabled Cyber–Physical Internet and systems analytics. His research has been supported by substantial government and industry grants exceeding HK\$100 million. He has led a strong research team and collaborated closely with leading academic and industrial organizations through joint projects and start-up companies. He has published extensively, and his work has been highly cited by the research community.

Dr. Huang serves as an Associate Editor and Editorial Board Member for several international journals. He is a Chartered Engineer and a Fellow of ASME, CILT, HKIE, IET, IEEE, and IISE.
\end{IEEEbiography}

\end{document}